\documentclass[11pt,letterpaper]{article}

\usepackage[
    top=1in,
    bottom=1in,
    left=1in,
    right=1in
]{geometry}

\usepackage[T1]{fontenc}
\usepackage[utf8]{inputenc}

\usepackage{newtxtext}

\usepackage{amsmath,amssymb,amsthm,mathtools}
\usepackage{newtxmath}
\usepackage{bm}

\usepackage{microtype}
\usepackage{xspace}

\usepackage{graphicx}
\graphicspath{{figures/}}
\usepackage{booktabs}
\usepackage{multirow}
\usepackage{subcaption}

\usepackage{enumitem}
\usepackage{titlesec}
\usepackage{xcolor}
\definecolor{linkblue}{RGB}{25,70,130}

\setlist[itemize]{leftmargin=*,topsep=3pt,itemsep=1pt,parsep=0pt}
\setlist[enumerate]{leftmargin=*,topsep=3pt,itemsep=1pt,parsep=0pt}

\titleformat{\section}
  {\large\bfseries}
  {\thesection}
  {0.65em}
  {}
\titlespacing*{\section}{0pt}{1.6ex plus 0.4ex minus 0.2ex}{0.8ex}

\titleformat{\subsection}
  {\normalsize\bfseries}
  {\thesubsection}
  {0.6em}
  {}
\titlespacing*{\subsection}{0pt}{1.3ex plus 0.3ex minus 0.2ex}{0.5ex}

\titleformat{\subsubsection}
  {\normalsize\itshape}
  {\thesubsubsection}
  {0.55em}
  {}
\titlespacing*{\subsubsection}{0pt}{1.0ex plus 0.2ex minus 0.1ex}{0.3ex}

\usepackage[numbers,sort&compress]{natbib}

\usepackage[
    colorlinks=true,
    linkcolor=linkblue,
    citecolor=linkblue,
    urlcolor=linkblue,
    pdfauthor={Xingji Cui},
    pdftitle={AugRelNet: Relation-Augmented Dynamics for Structure Discovery and Forecasting from Limited Data}
]{hyperref}
\usepackage[nameinlink,capitalise,noabbrev]{cleveref}

\newtheorem{theorem}{Theorem}[section]
\newtheorem{proposition}[theorem]{Proposition}
\newtheorem{lemma}[theorem]{Lemma}
\newtheorem{corollary}[theorem]{Corollary}

\theoremstyle{definition}

\theoremstyle{remark}

\begin{document}
\begin{center}
{\LARGE\bfseries AugRelNet: Relation-Augmented Dynamics for Structure Discovery and Forecasting from Limited Data\par}
\vspace{0.75em}
{\normalsize Xingji Cui\par}
\vspace{0.2em}
{\small Xi'an Jiaotong University\par}
{\small \texttt{cuixingji@stu.xjtu.edu.cn}\par}
\end{center}
\vspace{0.6em}
\begin{center}
\begin{minipage}{0.92\textwidth}
\small
\begin{center}\textbf{Abstract}\end{center}
\vspace{-0.35em}

Complex dynamics often arise from persistent interactions whose effects vary with the system state. We introduce AugRelNet for short- and medium-horizon trajectory forecasting from limited data. A shared learnable relation matrix explicitly routes source signals to targets, while evolving local features \(u\) and a persistent response state \(\eta\) transform them into nonlinear, history-dependent responses. Feeding these responses back into the dynamics separates shared relational organization from its changing realization. The learned relations reflect dense coupling in Lotka–Volterra systems and develop structured, neighbour-preferential organization in locally interacting systems. On 40-dimensional Lorenz–96, AugRelNet uses about 4,000 training states, roughly an order of magnitude fewer than representative prior Lorenz–96 forecasting studies. Across five seeds, every true interaction ranks above every nonedge, with a mean absolute-weight contrast of approximately \(146{:}1\). Under the same data budget, eight-step RMSE is about \(23\)–\(59\times\) lower than MNO, DySLIM, DeepSkip, and AL-RNN, \(14\)–\(55\times\) lower than NRI and dense GraphODE, and 37\% lower than PySINDy with a matched quadratic dictionary. The same relational representation extends to grids by treating sites as variables and sharing relations across spatial offsets: in discrete (Game of Life), continuous (FitzHugh–Nagumo), and stochastic (Ising) dynamics, learned weights rank physical neighbours above other candidate offsets while supporting multi-step prediction. AugRelNet learns these relations through prediction losses and regularization, without structural labels or system-specific physical losses.


\end{minipage}
\end{center}
\vspace{1.0em}
\section{Introduction}

Learning a dynamical system requires identifying not only how it changes, but also what remains reusable across those changes. Across trajectories, the effects of interactions may vary with state and history, while the pathways through which variables influence one another can remain persistent. These recurring patterns are information in the observations: repeated predictive dependencies can reveal what is shared across changing dynamics. Such reusable organization is especially valuable with limited observations, since interaction patterns need not be rediscovered independently across states.

The challenge is to represent this organization without restricting the dynamics it supports. Graphs or coefficient matrices make interactions explicit, but often require structural assumptions or tie relations to a particular interaction law. Fully flexible neural dynamics can represent state-dependent effects, but the organization behind those effects may remain implicit. The goal is therefore not simply to learn a graph and then forecast, but to represent where interactions are organized separately from how they are realized.

We introduce \textbf{AugRelNet}, a relation-structured augmented neural dynamical model built around this separation. Cross-variable dependence is forced through an explicit relational channel $D$, rather than being hidden inside unconstrained latent transformations. The matrix $D$ specifies where relational influence can flow, while nonlinear dynamics determine how that influence is realized at the current state. Thus, $D$ represents persistent relational organization rather than a fixed interaction coefficient. This structure also extends beyond instantaneous effects: finite-time influence remains organized through $D$-weighted relational paths while the underlying dynamics remain state dependent.

The augmented dynamics assign separate roles to local capacity and persistent responses. The auxiliary state $u$ provides flexible local predictive capacity, whereas $\eta$ retains responses generated through the relational pathway over learned time scales. Together, the relation channel and augmented states separate where interactions are organized, how they are realized, and how their effects persist over time. When known spatial symmetries are available, the same relational representation can share coordinates across spatial offsets while leaving active interactions learned. All components are optimized jointly from prediction losses and regularization, without structural labels or system-specific physical objectives.

On 40-dimensional Lorenz--96, AugRelNet learns among all candidate directed interactions from only 3,849 training states without the cyclic topology or an interaction mask. Across five seeds, every true interaction ranks above every nonedge, with an edge-to-nonedge contrast of approximately $146{:}1$. Under the same low-data protocol, eight-step RMSE is reduced by about $14$--$59\times$ relative to tested forecasting and relation-learning baselines without graph or equation priors. At short horizons, mean RMSE is also below true-graph GraphODE-Oracle and matched-quadratic PySINDy. Relation interventions and capacity-matched ablations support the roles of learned routing and structured response states.

The learned relations are not tied to a single physical notion of structure. In generalized Lotka--Volterra systems, they follow dense signed interaction geometry; in Lorenz--96, they recover sparse directed support. Under coarser observations, learned relations broaden consistently with finite-time predictive pathways. On FitzHugh--Nagumo, Game of Life, and stochastic Ising lattices, shared spatial relations identify physically consistent neighbourhoods across continuous, discrete, and probabilistic dynamics. Across these systems, prediction reveals reusable relational organization, while augmented dynamics determine how that organization is realized over state and time.
\section{Related Work}
\label{sec:related_work}
\noindent\textbf{Learning dynamical systems.} Trajectory-based modeling spans equation discovery \citep{brunton2016sindy}, reservoir and recurrent predictors \citep{platt2022systematic,brenner2024almost}, neural vector fields \citep{chen2018neural}, and neural operators \citep{li2021fno}. For chaotic systems, MNO learns solution operators for dissipative dynamics \citep{li2022chaotic} and DySLIM regularizes invariant measures \citep{schiff2024dyslim}; recent work also studies deep random features, attention-based forecasting, and pretrained dynamical models \citep{mandal2025random,he2025chaos,lai2026panda}.

\noindent\textbf{Learning relational dynamics.} Graph simulators propagate over supplied or constructed connectivity \citep{battaglia2016interaction,sanchez2020learning}, whereas NRI jointly infers latent interactions and predictive dynamics \citep{kipf2018nri}. Later work extends relation learning to continuous time, trainable adjacency, and learned structural priors \citep{zhang2024nrdi,kang2024online,gong2025ipsi,shao2026diffprior}. Prediction-based neighbourhood selection provides a related route \citep{li2024hogrc,braun2026causal}, while recent theory studies identifiability under dependent inputs and support geometry \citep{kuskova2026interaction}.

\noindent\textbf{Augmentation, memory, and local dynamics.} Augmented Neural ODEs enlarge the evolving state \citep{dupont2019augmented}, and MemDNRI adds entity and relation memories to relational inference \citep{gong2021memory}. Neural cellular automata and recurrent variants similarly combine hidden state with shared local updates on spatial systems \citep{richardson2024learning,kang2024locally}. AugRelNet instead couples explicit persistent routing with evolving augmented states, using variable-pair or spatial-offset relations within the same organization--realization decomposition.
\section{Relation-Structured Augmented Neural Dynamics}
\label{sec:method}

Observed variables retain their identity, each augmented with predictive coordinates. Cross-variable dependencies pass through an explicit communication channel whose weights are learned through prediction; expressive local dynamics determine the responses along those routes.

\subsection{Shared Relations and Local State}
\label{sec:relation_pathway}

For component $i$, let
\begin{equation}
q_i=[z_i,u_i,\eta_i],
\qquad D\in\mathbb R^{N\times N},\quad D_{ii}=0.
\label{eq:augmented_state}
\end{equation}
$D_{ij}$ weights source $i$ to target $j$; column $j$ contains incoming relations. It is a continuous signed parameter shared across samples and held constant during evolution. Self-effects belong to the local dynamics. For multivariate systems, every non-self route is a candidate.

Shared source maps produce a scalar signal and a vector message,
\begin{equation}
s_i=\mathcal S_\theta(q_i),\qquad
m_i=\mathcal M_\theta(q_i).
\end{equation}
The incoming representation is
\begin{equation}
\rho_j=\sum_{i\ne j}D_{ij}\Psi_\theta(q_i),
\qquad \Psi_\theta(q_i)=[s_i,s_i m_i].
\label{eq:relation_aggregation}
\end{equation}
The receiver combines $\rho_j$ with $q_j$ and local history context $b_j$. For a smooth local field $V_j(q_j,b_j,\rho_j)$,
\begin{equation}
\partial_{q_i}V_j
=D_{ij}\,\partial_{\rho_j}V_j\,\partial_{q_i}\Psi_\theta,
\qquad i\ne j.
\label{eq:method_cross_response}
\end{equation}
Thus $D$ specifies routing, not the physical Jacobian: realized sensitivities can change magnitude or sign with $D$ fixed. Nonlinear receivers can combine routed sources, allowing higher-order interactions despite linear aggregation; the mixed-response identity is given in Appendix~\ref{app:directed_propagation_proof}.

\subsection{Augmentation and Persistent Response}
\label{sec:augmented_dynamics}

Shared encoders map each component's causal history $\mathcal H_i(t_0)$ of observations and increments to auxiliary states and context, while inserting the observed coordinate directly:
\begin{equation}
\begin{aligned}
[u_i(t_0),\eta_i(t_0)]&=\mathcal I_\theta(\mathcal H_i(t_0)),\\
b_i&=\mathcal B_\theta(\mathcal H_i(t_0)),\\
q_i(t_0)&=[z_i(t_0),u_i(t_0),\eta_i(t_0)].
\end{aligned}
\label{eq:history_initialization}
\end{equation}
The componentwise encoders provide no communication path bypassing $D$. By default, $b_i$ stays fixed and forecasts use $\widehat z_i(t)=\Pi_zq_i(t)$, without a learned global decoder. The auxiliaries evolve during rollout rather than merely widening the instantaneous network.

Write $x_i=[z_i,u_i]$. For a neural map $\mathcal K$, define its zero-anchored relational response by
\begin{equation}
\begin{split}
\Delta_\rho\mathcal K(q_i,b_i,\rho_i)
={}&\mathcal K(q_i,b_i,\rho_i)\\
&-\mathcal K(q_i,b_i,0).
\end{split}
\label{eq:zero_anchored_response}
\end{equation}
The multivariate continuous-time realization is
\begin{align}
\dot x_i
&=\mathcal F^{\rm self}_\theta(q_i,b_i)
 +\Delta_\rho\mathcal F^{\rm rel}_\theta(q_i,b_i,\rho_i)
 +d_i e_z,
\label{eq:augmented_x}\\
\dot\eta_i&=\Lambda_i\bigl[g_i(q_i,b_i,\rho_i)-\eta_i\bigr],
\label{eq:augmented_eta}\\
g_i&=\Delta_\rho\mathcal G_\theta(q_i,b_i,\rho_i),
\qquad \Lambda_i=\operatorname{diag}(\lambda_i),\quad\lambda_i>0.
\label{eq:augmented_drive}
\end{align}
Here $d_i$ is a learned scalar drift, $e_z$ selects the observed coordinate of $x_i$, and positive rates are parameterized by softplus. Neural maps are shared; states, contexts, relation weights, drifts, and relaxation rates allow component-specific behavior.

The roles of $u$ and $\eta$ differ. $u$ evolves freely; after initialization $\eta$ is driven only by the zero-anchored relational response $g_i$, so $g_i(q_i,b_i,0)=0$, and retains it over learned time scales:
\begin{equation}
\begin{split}
\eta_i(t)={}&e^{-\Lambda_i(t-t_0)}\eta_i(t_0)\\
&+\int_{t_0}^{t}e^{-\Lambda_i(t-s)}\Lambda_i g_i(q_i(s),b_i,\rho_i(s))
\,\mathrm ds .
\end{split}
\label{eq:eta_memory}
\end{equation}
Thus $\eta$ carries temporal information through a response state tied to the relational channel rather than an unrestricted recurrent state. Unlike fixed $b$, it accumulates responses during rollout; rolling-$b$ and resetting $\eta$ test different mechanisms. Section~\ref{sec:lattice_specialization} retains this routing--augmentation principle with task-specific updates.

\subsection{Prediction Training}
\label{sec:training}

Models are trained on causal history--future windows, using weighted forecast losses of the form
\begin{equation}
\mathcal L_{\rm pred}
=\mathbb E_{\rm windows}\sum_{h\in\mathcal H}
w_h\,\ell(\widehat z_{t+h},z_{t+h}),\qquad w_h\ge0,
\label{eq:prediction_loss}
\end{equation}
together with benchmark-specific regularization. Relations and response maps are optimized jointly from prediction, without structural labels, derivative targets, invariant measures, or system-specific physical equations. Continuous benchmarks use Heun integration, holding $D$ fixed while recomputing emissions, relational aggregates, and nonlinear responses at both stages. Thus relational organization persists across a step while its state-dependent realization is reevaluated at the predictor and corrector states. Full losses and training protocols are given in Appendix~\ref{app:protocol}.

\subsection{Properties of the Relational Dynamics}
\label{sec:relational_properties}

The following results distinguish propagation in a fitted field, selection among prediction-equivalent parameters, and the effect of the response state on prediction errors. The continuous-time results use the smooth, finite-horizon setting of Appendix~\ref{app:relational_setup}; the selection result concerns a frozen predictive tangent model.

\begin{proposition}[Directed relational propagation]
\label{prop:directed_relational_propagation}
For fixed $D$ and conditioners, with all cross-node dependence passing through Eq.~\eqref{eq:relation_aggregation},
\begin{equation}
\begin{aligned}
\frac{\partial q_j(t)}{\partial q_i(0)}
&=\mathbf 1_{i=j}U_j(t,0)
 +\sum_{r\ge1}\sum_{p\in\mathcal W^r_{i\to j}}D_p\mathcal K_p(t),\\
\|\mathcal K_p(t)\|&\le e^{Lt}(Mt)^r/r! .
\end{aligned}
\label{eq:shared_walks}
\end{equation}
Here $\mathcal W^r_{i\to j}$ contains length-$r$ directed walks, $D_p$ is the product of their relation weights, and $U_j$ is the node-local variational propagator. The kernels depend on the evolving states, and $L,M$ are uniform bounds specified in Appendix~\ref{app:directed_propagation_proof}. A length-$r$ walk contributes at order $t^r$ or higher; cancellation can remove its leading term.
\end{proposition}

\paragraph{Selection under non-identifiability.} For fixed forecast windows, let $J_D\in\mathbb R^{n\times m}$ and $J_\theta\in\mathbb R^{n\times p}$ be frozen prediction Jacobians for relation coordinates $d$ and other parameter coordinates $w$. Assume $J_D$ has full column rank and the target displacement is $J_Da$. Define the compensating geometry
\begin{equation}
K=J_D^\dagger J_\theta,\qquad
R=(I-J_DJ_D^\dagger)J_\theta,\qquad
C=K\Pi_{\ker R}K^{\mathsf T},
\label{eq:compensation_geometry}
\end{equation}
where $\Pi_{\ker R}$ is the orthogonal projector onto directions whose prediction changes can be offset through $d$.

\begin{proposition}[Gradient-selected relational coordinates]
\label{prop:relational_selection}
Unregularized Euclidean gradient flow from $d=w=0$ on $\frac12\|J_Dd+J_\theta w-J_Da\|_2^2$ selects
\begin{equation}
d_*=(I+C)^{-1}a.
\label{eq:relational_selection}
\end{equation}
The selected map preserves every coordinate support if and only if $C$ is diagonal, and every coefficient direction if and only if $C=\kappa I$ for some $\kappa\ge0$.
\end{proposition}

Thus gradient selection can preserve sparse or dense organization even when its parameter allocation is non-unique. The decisive property is whether compensation mixes relation coordinates, not a prescribed sparsity pattern. Appendix~\ref{app:selection_proof} gives the proof, initialization scope, and a perturbation bound for strict edge ranking; Corollary~\ref{thm:implicit_D_alignment} specializes the result to GLV.

\paragraph{Relaxation-weighted response errors.} Fix $D,b,\Lambda$ and write the continuous realization as $\dot X=F_D(X,E)$, $\dot E=\Lambda(G_D(X,E)-E)$, with $X=(z,u)$ and $E=\eta$. Define
\begin{equation}
(\mathcal T_\Lambda e)(t)
=\int_0^t e^{-\Lambda(t-s)}\Lambda e(s)\,\mathrm ds,
\qquad
\|v\|_{\infty,T}=\sup_{0\le t\le T}\|v(t)\|_\infty .
\label{eq:response_error_filter}
\end{equation}

\begin{proposition}[Relaxation-weighted response approximation]
\label{prop:response_approximation}
Suppose $F_D,G_D$ are Lipschitz on a common domain containing the compared trajectories. Starting from the same full state, perturb only the response drive to $G_D(\widetilde X,\widetilde E)+e(t)$ for a bounded error $e$. Then
\begin{equation}
\|\widetilde X-X\|_{\infty,T}
\le C_T\|\mathcal T_\Lambda e\|_{\infty,T},
\label{eq:filtered_forecast_error}
\end{equation}
where $C_T$ depends on the common Lipschitz bounds, rates, and horizon, but not on $e$.
\end{proposition}

The bound retains temporal cancellation in response errors instead of depending only on their instantaneous amplitude, including for Markovian target dynamics. Appendix~\ref{app:response_approximation} gives the closed-loop proof and time-scale trade-off, and distinguishes this continuous-time result from the implemented Heun updates.

\section{Multivariate Dynamics}
\label{sec:multivariate}

We examine both dense interactions and local chaotic coupling. Forecast accuracy measures the learned evolution, while matrix comparisons and interventions examine how the model represents and uses relations. Unless stated otherwise, reported dispersions are population standard deviations across seeds. The GLV seeds generate different physical systems; Lorenz--96 seeds vary the learned model on one fixed dataset.

\subsection{Lotka--Volterra: Dense Interaction Geometry}
\label{sec:glv}

Generalized Lotka--Volterra (GLV) dynamics are
\begin{equation}
\dot x_i=x_i\left(r_i+\sum_j A_{ij}x_j\right),
\label{eq:glv}
\end{equation}
where $A_{ij}$ is the effect of source $j$ on target $i$. Its off-diagonal entries provide a dense reference for $D^{\mathsf T}$. The physical cross-derivative is instead $x_iA_{ij}$.

AugRelNet improves over Neural ODE at all three horizons (Table~\ref{tab:glv_prediction}), with a mean horizon-wise RMSE reduction of $74.5\%$. In four of five systems, the absolute off-diagonal Pearson correlation between $D^{\mathsf T}$ and $A$ is $0.9960\pm0.0008$. The fifth has correlation $0.767$ but retains $R^2=0.995$ at the longest horizon, with discrepancies concentrated in weak interactions (Appendix~\ref{app:glv_diagnostics}). We compare relative signed geometry up to global orientation and scale.

\begin{figure}[htbp]
\centering
\begin{minipage}{0.42\linewidth}
\centering
\includegraphics[width=0.85\linewidth]{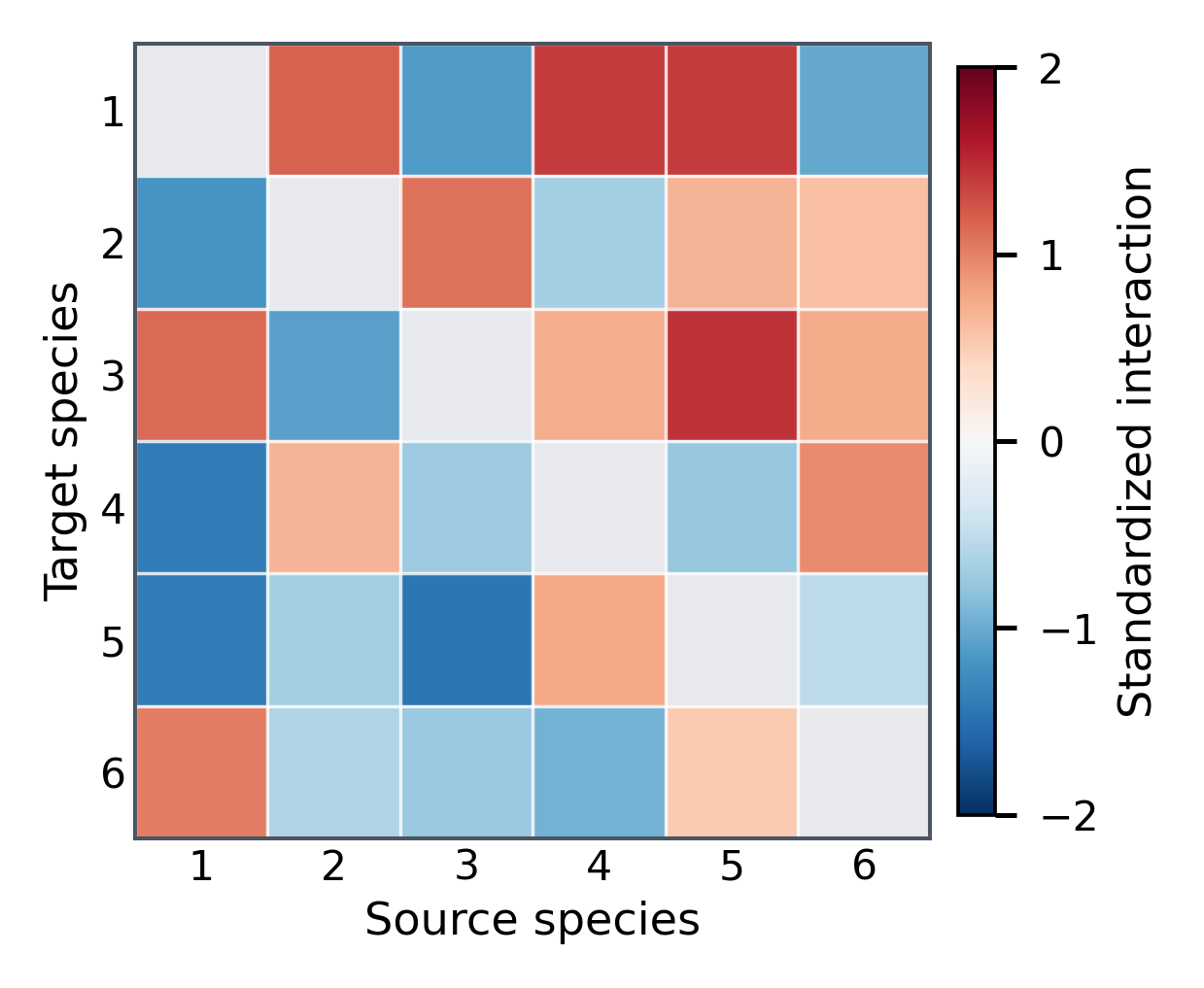}\\
{Ground truth $A$}
\end{minipage}\hspace{0.04\linewidth}
\begin{minipage}{0.42\linewidth}
\centering
\includegraphics[width=0.85\linewidth]{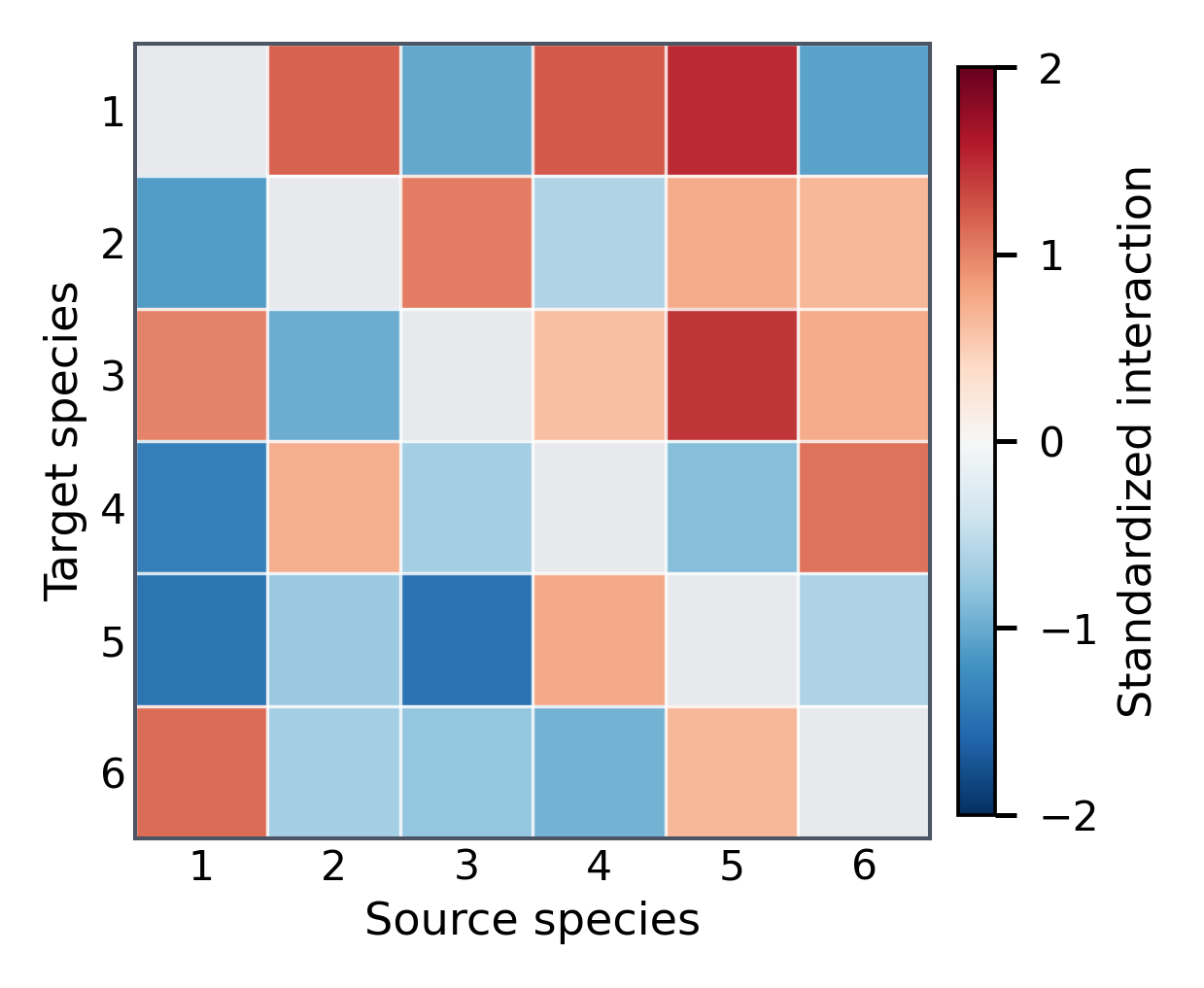}\\
{Learned $D^{\mathsf T}$}
\end{minipage}
\caption{GLV seed 0: targets in rows, sources in columns, diagonal
masked. Each matrix is separately standardized off-diagonal;
the shared colour scale compares relative organization, not raw magnitudes.}
\label{fig:glv_D_A_alignment}
\end{figure}

\paragraph{Selection despite non-identifiability.} The shared functional form of GLV permits a coefficient-level specialization of Proposition~\ref{prop:relational_selection}. Let $a=\operatorname{vec}(A_{\rm off})$ and $d=\operatorname{vec}(D^{\mathsf T}_{\rm off})$. In a frozen local prediction model, write $\widehat y=\Phi(cd+Kw)$ and $y=\Phi a+\epsilon$, with shared gain $c\ne0$ and $\Phi^{\mathsf T}\Phi\succ0$. The discrepancy $\epsilon$ includes finite-time and model-approximation errors.

\begin{corollary}[GLV coefficient alignment]
\label{thm:implicit_D_alignment}
Euclidean gradient flow from zero on $\frac12\|\Phi(cd+Kw)-y\|_2^2$ selects
\begin{equation}
d_*=c(c^2I+B)^{-1}(a+\delta a),\qquad
B=KK^{\mathsf T},\quad \delta a=\Phi^\dagger\epsilon.
\label{eq:local_allocation}
\end{equation}
When $\epsilon=0$, the direction of a nonzero $a$ is preserved exactly if and only if $Ba=\kappa_a a$; then $d_*=c(c^2+\kappa_a)^{-1}a$. Thus global isotropy of the compensation channel is sufficient but unnecessary for alignment of a particular system.
\end{corollary}

Misalignment is controlled by the component of $Ba$ orthogonal to $a$ and by $\delta a$. Appendix~\ref{app:glv_alignment_proof} derives the corresponding Pearson-correlation and signed-weight bounds. These explain conditions under which strong coefficients remain aligned while weak ones deviate; the measured alignment remains an empirical outcome of training.

\subsection{Lorenz--96: Learning Relations from Limited Data}
\label{sec:l96}

Lorenz--96 combines high-dimensional chaos with a repeated local law,
\begin{equation}
\dot x_j=(x_{j+1}-x_{j-2})x_{j-1}-x_j+F,
\label{eq:l96_system}
\end{equation}
with cyclic indexing. We use $N=40$, $F=8$, and 3,849 training states at $\Delta=0.02$, roughly an order of magnitude fewer than representative 40-dimensional forecasting studies \citep{platt2022systematic,mandal2025random}. All controlled comparisons use this low-data budget. AugRelNet learns a dense relation matrix from prediction without the cyclic topology, an interaction mask, or structural supervision.

\begin{table}[!t]
\centering
\setlength{\tabcolsep}{4pt}
\caption{Lorenz--96 physical RMSE on 785 common origins under the shared
low-data budget; mean $\pm$ population standard deviation over five model
seeds (PySINDy deterministic). Bold: lowest mean without a supplied true graph
or matched equation class.}
\label{tab:l96_baseline_comparison}
\begin{tabular}{lcccc}
\toprule
Method & $h=1$ & $h=8$ & $h=24$ & $h=40$ \\
\midrule
\multicolumn{5}{l}{\textit{Dynamical-system models}} \\
MNO & $0.768\pm0.047$ & $2.675\pm0.057$ & $4.362\pm0.117$ & $4.582\pm0.069$ \\
DySLIM & $0.394\pm0.007$ & $2.919\pm0.061$ & $8.301\pm0.188$ & NF (4/5) \\
Global DeepSkip & $0.151\pm0.003$ & $1.126\pm0.027$ & $3.627\pm0.061$ & $14.918\pm18.298^{\dagger}$ \\
AL-RNN & $0.642\pm0.022$ & $2.449\pm0.015$ & $3.512\pm0.055$ & $3.687\pm0.033$ \\
\midrule
\multicolumn{5}{l}{\textit{Relation-aware models}} \\
NRI & $0.114\pm0.012$ & $0.710\pm0.056$ & $3.888\pm0.118$ & $5.082\pm0.100$ \\
GraphODE-Dense & $0.378\pm0.001$ & $2.711\pm0.010$ & $4.857\pm0.043$ & $5.014\pm0.065$ \\
\midrule
AugRelNet & $\mathbf{0.0061}\pm0.0003$ & $\mathbf{0.049}\pm0.003$
& $\mathbf{0.250}\pm0.016$ & $\mathbf{0.685}\pm0.051$ \\
\midrule
\multicolumn{5}{l}{\textit{Strong-prior references}} \\
GraphODE-Oracle & $0.007\pm0.001$ & $0.055\pm0.007$ & $0.203\pm0.039$ & $0.451\pm0.083$ \\
PySINDy poly2 & $0.010$ & $0.078$ & $0.271$ & $0.576$ \\
\bottomrule
\end{tabular}
\vspace{3pt}
\begin{minipage}{0.97\textwidth}
NF: four DySLIM seeds are nonfinite at $h=40$.
$^\dagger$ One DeepSkip run reaches $51.514$ at $h=40$ and is retained.
Oracle receives the true 120 directed edges; poly2 uses the matched
quadratic dictionary.
\end{minipage}
\end{table}

\begin{figure}[t]
\centering
\includegraphics[width=0.90\linewidth]{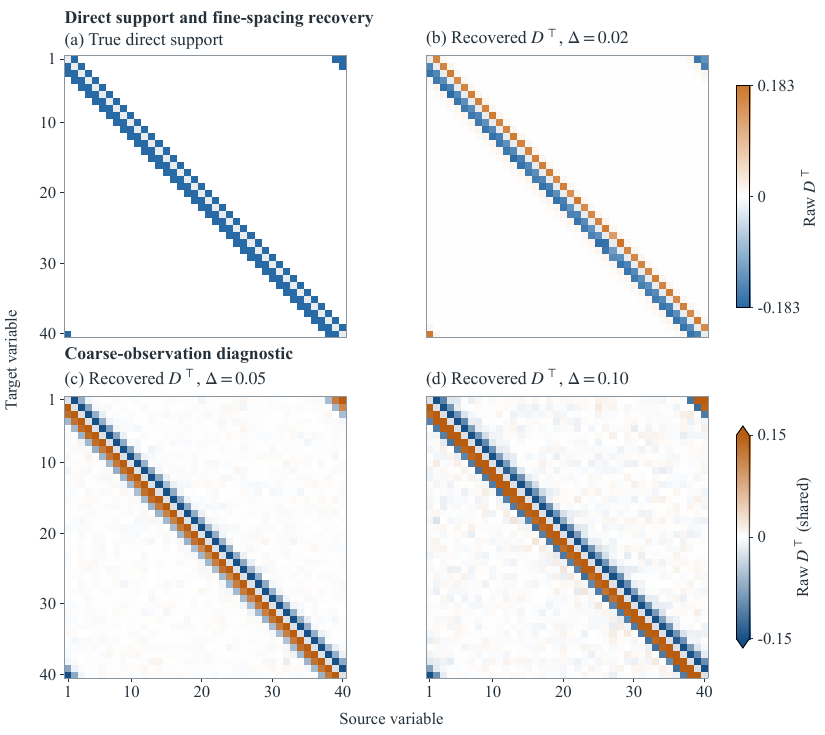}
\caption{\textbf{Direct support recovery and coarser-observation broadening.}
(a) True support; (b) learned signed $D^{\mathsf T}$ at $\Delta=0.02$;
(c,d) the direct core persists at $\Delta=0.05,0.10$ while off-support weight
broadens along graph distance. Separate raw-weight scales are used for (b) and
(c,d). Seed 0. }
\label{fig:l96_relations_spacing}
\end{figure}

\noindent\textbf{Forecasting.}\ All models share the chronological split, 785 held-out origins, and validation-based selection. Fixed-$b$ AugRelNet leads models without a supplied true graph or matched equation class at every horizon (Table~\ref{tab:l96_baseline_comparison}). At $h=8$, RMSE is about $23$--$59$ times lower than MNO, DySLIM, DeepSkip, and AL-RNN and $14$--$55$ times lower than NRI and dense GraphODE. At $h=8$, RMSE is also 37\% below matched-dictionary PySINDy without that prior; PySINDy leads at $h=40$.

\noindent\textbf{Recovered relations.}\ Across all five seeds, the globally largest $3N=120$ of 1,560 off-diagonal $|D|$ entries coincide exactly with the true support: every true edge exceeds every nonedge (Fig.~\ref{fig:l96_relations_spacing}(a,b)). The $j+1$ and $j-2$ weights have opposite signs at all 40 targets, matching Eq.~\eqref{eq:l96_system}, and the mean edge-to-nonedge absolute-weight contrast is about $146{:}1$. Prediction training thus recovers sharply separated, approximately sparse organization without fixing three edges per target. At the same 120-edge budget, the NRI adaptation recovers 31, 44, 9, 54, and 45 true edges in its five seeds; unlike $D$, these graphs are history-conditioned.

\noindent\textbf{Coarser observations.}\ Keeping architecture and observation count fixed, increasing $\Delta$ from $0.05$ to $0.10$ raises the two-hop shell's share of off-diagonal squared weight from $3.3\%$ to $10.1\%$; across all six runs, this shell contains $90$--$96\%$ of nonedge weight (Fig.~\ref{fig:l96_relations_spacing}(c,d)). The direct core persists while off-support mass broadens along graph distance, separating \emph{instantaneous} from \emph{finite-time predictive} locality in a manner consistent with Proposition~\ref{prop:directed_relational_propagation}.

\noindent\textbf{Do these relations support prediction?}\ On the initial checkpoints, zeroing or shuffling $D$ raises $h=8$ standardized RMSE by $26.0\times$ and $29.8\times$ across five seeds. Resetting $\eta$ after each completed Heun step raises seed-0 $h=8$ error by $15.0\times$ while leaving the first prediction unchanged, as characterized by Lemma~\ref{prop:persistent_response_reset}.

\noindent\textbf{Response-state structure matters beyond capacity.}\ Capacity-matched variants retain the same relation pathway and conditioner. At $h=8$, RMSE is $.083$, $.107$, $.131$, and $.367$ for $(z,u,\eta)$, $(z,\eta)$, $(z,u)$, and $z$, respectively. In particular, $(z,\eta)$ outperforms the free-augmentation control $(z,u)$, with the full state best (Table~\ref{tab:l96_state_ablations}).

Strict support separation also persists when direct regularization on $D$ is removed in an additional seed-0 run (Appendix~\ref{app:l96_additional_diagnostics}).

\begin{figure}[!t]
    \centering
    \includegraphics[width=0.90\linewidth]{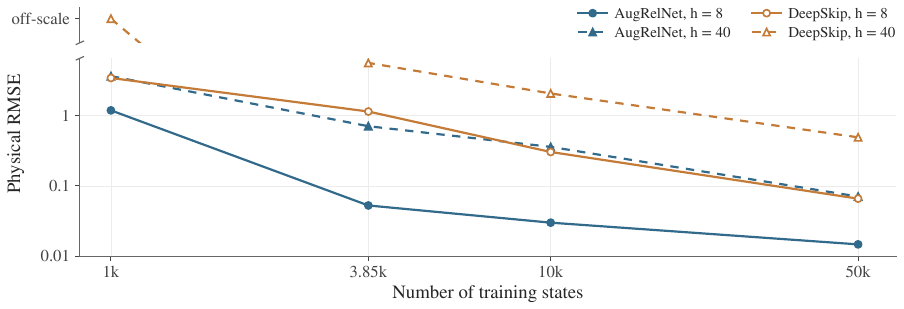}
    \caption{\textbf{Lorenz--96 data scaling.} Physical RMSE at $h=8,40$ across
    four training budgets; DeepSkip's 1k-state $h=40$ value ($4196.22$) is
    shown off-scale.}
    \label{fig:l96_scaling}
\end{figure}

\noindent\textbf{Data scaling.}\ Across a controlled single-seed extension from 1,000 to 50,000 training states, AugRelNet has lower physical RMSE than DeepSkip at every evaluated horizon (Fig.~\ref{fig:l96_scaling}); at 50k the gap remains $4.46\times$ at $h=8$ and $6.95\times$ at $h=40$. Exact top-120 support recovery is already reached at 3,849 states, while forecast error continues to improve (Appendix~\ref{app:l96_scaling_longrun}).

\noindent\textbf{Long-horizon behavior.}\ DeepSkip more closely reproduces the reference long-run statistics, consistent with AugRelNet not being optimized for invariant-statistics fidelity. Still, a 50k-trained AugRelNet remains stable for 50,000 autonomous steps from all five starts without resets or clamp activation and retains organized space--time structure, despite shifted mean energy ($12.68$ versus $9.36$); full diagnostics are in Appendix~\ref{app:l96_scaling_longrun}.

\section{Learning Spatial Relations on Lattices}
\label{sec:lattice_specialization}

\begin{figure}[!t]
    \centering
    \includegraphics[width=0.90\textwidth]{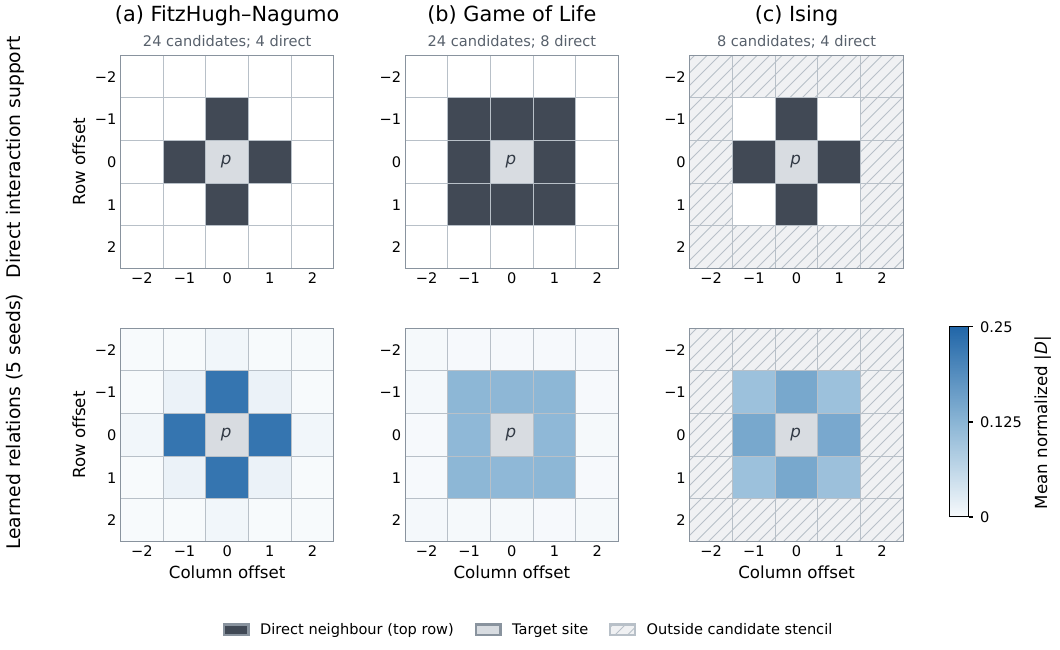}
    \caption{Spatial support (top) and mean normalized learned $|D|$ over five
    seeds (bottom). All seeds rank the physical neighbours above other candidate
    offsets; hatching marks locations outside the candidate stencil.}
    \label{fig:lattice_neighbourhoods}
\end{figure}

We next treat each grid site $p$ as a dynamical variable, retaining its local state $q_p^n=[\bar z_p^n,u_p^n,\eta_p^n]$. Relation weights are shared by spatial offset:
\begin{equation}
D_{(p+\delta),p}=D_\delta,\qquad \delta\in\mathcal N,
\label{eq:lattice_shared_relations}
\end{equation}
where $\mathcal N$ is a finite candidate stencil. This factors out known translation symmetry rather than relearning it at every site: the symmetry and candidate family are supplied, while active offsets are learned (Fig.~\ref{fig:lattice_neighbourhoods}).

\noindent\textbf{From variable pairs to spatial offsets.}\ With boundary-validity mask $\chi_{p,\delta}$, the shared routing aggregates
\begin{equation}
\rho_p=c_{\mathcal N}\sum_{\delta\in\mathcal N}
\chi_{p,\delta}D_\delta
\Psi_\theta(q_{p+\delta},e_\delta),
\label{eq:lattice_aggregation}
\end{equation}
so relation parameters scale with candidate offsets rather than site pairs. FHN and Life use 24 radius-two offsets and Ising eight; details are in Appendix~\ref{app:protocol}.

\noindent\textbf{Predictive state and recorded-step evolution.}\ Continuous fields retain their visible coordinate, whereas Life and Ising write predicted occupancy probabilities or expected spins back autoregressively. Auxiliaries evolve alongside these predictions; writeback details are in Appendix~\ref{app:lattice_writeback}. Ising forecasts are sitewise probabilities, not sampled joint transitions. Table~\ref{tab:lattice_forecasts} compares common origins.

\begin{table}[!t]
\centering
\setlength{\tabcolsep}{2.3pt}
\caption{Lattice forecasts on common test origins within each system;
mean $\pm$ population standard deviation over five seeds where applicable.
FHN RMSE is standardized. Lower is better. FHN RMSE and Life BCE values are multiplied by $10^3$ for display.}
\label{tab:lattice_forecasts}
\begin{tabular}{llccc}
\toprule
System / metric & Model & $h=1$ & $h=4$ & $h=8$\\
\midrule
\multirow{3}{*}{\shortstack[l]{FHN\\RMSE $\times10^3$}}
& FNO & $8.685\pm2.747$ & $36.201\pm9.779$ & $61.386\pm14.473$\\
& U-Net & $8.195\pm.643$ & $15.085\pm.968$ & $15.163\pm1.162$\\
& AugRelNet & $\mathbf{.241}\pm.042$ & $\mathbf{1.327}\pm.221$ & $\mathbf{1.614}\pm.271$\\
\midrule
\multirow{3}{*}{\shortstack[l]{Life\\BCE $\times10^3$}}
& CNN $5\times5$ & $4.292\pm.978$ & $39.831\pm5.863$ & $72.531\pm5.235$\\
& AugRelNet & $.006\pm.011$ & $.196\pm.290$ & $.894\pm1.228$\\
& PolyKAN $3\times3$ ref. & $.001\pm.000$ & $.001\pm.000$ & $.001\pm.000$\\
\midrule
\multirow{4}{*}{\shortstack[l]{Ising\\BCE}}
& Kinetic local logistic & $.28885\pm.00011$ & $.35674\pm.00002$ & $.38736\pm.00019$\\
& CNN $5\times5$ & $\mathbf{.28507}\pm.00030$ & $.35514\pm.00036$ & $\mathbf{.38131}\pm.00013$\\
& AugRelNet & $.28520\pm.00003$ & $\mathbf{.35467}\pm.00003$ & $.38134\pm.00012$\\
& Oracle & $0.28348$ & $-$ & $-$\\
\bottomrule
\end{tabular}
\vspace{2pt}
\begin{minipage}{.97\textwidth}
PolyKAN has a prescribed $3\times3$ receptive field and reaches the BCE
clipping floor. Bold: lowest FHN/Ising mean. Life uses the main test set; full
evaluation details are in Appendix~\ref{app:protocol}.
\end{minipage}
\end{table}

\subsection{FitzHugh--Nagumo: Continuous Fields}
\label{sec:lattice_fhn}

On a $64\times64$ FitzHugh--Nagumo grid, AugRelNet learns among 24 candidate offsets containing the four physical diffusion neighbours. Across five seeds it reaches $0.00189\pm0.00022$ standardized RMSE at $h=16$ (U-Net: $0.01700$; FNO: $0.08892$), ranks the four neighbours highest ($84$--$93\%$ of total $|D|$), and incurs about $114\times$ higher error when their assignments are permuted. Beyond training, $h=32$ RMSE rises to $0.256$ (U-Net: $0.0226$). Full protocols and model sizes are in Appendix~\ref{app:protocol}.

\subsection{Game of Life: Discrete Neighbourhood Rules}
\label{sec:lattice_life}

On periodic $64\times64$ grids, AugRelNet learns among 24 candidate offsets rather than receiving the Moore stencil. Four-step exact-grid accuracy (all 4,096 cells) is $98.1\%$ (PolyKAN: $100\%$) and $95.7\pm8.0\%$ on fresh initial conditions without retraining. All five seeds rank exactly the eight Moore neighbours highest; retaining only them preserves $98.1\%$ accuracy, whereas retaining only the outer sixteen gives zero. Thus the dominant routes suffice while the nonlinear local update implements their joint effect.

\subsection{Ising: Probabilistic Forecasting}
\label{sec:ising}

For checkerboard heat-bath Ising at $\beta\approx0.44069$, the exact full-sweep one-step oracle attains BCE $0.28348$, versus $0.28520$ for AugRelNet and $0.28507$ for the CNN; at $h=8$, the models remain matched (Table~\ref{tab:lattice_forecasts}). Under single-site Gibbs resampling, correlations give non-neighbours marginal predictive information (Proposition~\ref{prop:ising_predictive_dependence}); a recorded sweep additionally composes updates through neighbours. Diagonal routes can therefore carry predictive information without direct physical interactions. All five seeds rank the cardinal neighbours above the diagonals (mean ratio $1.41$). Swapping these groups raises expected H1 excess BCE above the oracle by $117.6\%$; all 40 random model--permutation pairs increase excess loss, whereas within-group permutations have negligible effect. Zeroing $D$ raises $h=8$ BCE from approximately $0.381$ to $0.699$ (Appendix~\ref{app:ising_diagnostics}).

\section{Limitations and Scope}

Unrestricted directed relation discovery uses $O(N^2)$ coordinates; reducing this cost requires structural assumptions or sharing. Saturated forecast accuracy can also require low learning rates and long optimization, although useful structure and stable prediction often emerge earlier. $D$ represents predictive organization at the observation scale rather than a universally identifiable physical or causal relation; when this organization is weak, AugRelNet may approach a generic learned mapping. Finite-horizon accuracy does not ensure long-run fidelity: Lorenz--96 remains stable but matches long-run statistics less closely than DeepSkip, while FHN degrades beyond supervision. Partial observation further requires hidden pathways to be represented through effective observed-variable relations, making relation learning more demanding.
\clearpage
\section*{AI Use Statement}

Generative AI tools were used for research discussion, mathematical derivations and proof organization, experimental and coding assistance including debugging and code organization, literature search, result interpretation, and manuscript editing. All mathematical claims, citations, experiments, and reported results were independently checked by the author, who takes responsibility for the final paper.

\section*{Code Availability}

Code, configurations, selected checkpoints, and companion artifacts are available at
\url{https://github.com/Cuixjnoob/AugRelNet}.
\bibliographystyle{plainnat}
\bibliography{references}

\clearpage

\appendix

\section{Relational Selection and GLV Alignment}
\label{app:implicit_D_alignment}
\label{app:selection_proof}

\subsection{Predictive coordinates and gradient selection}
\label{app:selection_setup}
For fixed forecast windows, subtract a reference prediction and freeze its Jacobians $J_D\in\mathbb R^{n\times m}$ and $J_\theta\in\mathbb R^{n\times p}$ with respect to relation and remaining parameter coordinates. The local predictor is $J_Dd+J_\theta w$. Coordinates of $d$ retain their source--target indexing: no rotation or whitening is used to obtain a desired support. We assume $J_D$ has full column rank and first consider the compatible target $J_Da$. Thus the target coordinate $a$ is determined within this tangent subspace, although its allocation between $d$ and $w$ need not be unique. The loss in Proposition~\ref{prop:relational_selection} has no explicit regularizer.

Write $P_D=J_DJ_D^\dagger$, $K=J_D^\dagger J_\theta$, $R=(I-P_D)J_\theta$, and $\Pi=\Pi_{\ker R}$. Then $J_\theta=J_DK+R$ and $\operatorname{range}R$ is orthogonal to $\operatorname{range}J_D$. Only directions in $\ker R$ can compensate for a relation change without also changing the prediction outside the relational tangent subspace.

\begin{proof}[Proof of Proposition~\ref{prop:relational_selection}]
Set $L=[J_D\ J_\theta]$ and $z=(d,w)$. With optimization time $s$, the residual $r=Lz-J_Da$ satisfies
\begin{equation}
\frac{\mathrm dz}{\mathrm ds}=-L^{\mathsf T}r,\qquad
\frac{\mathrm dr}{\mathrm ds}=-LL^{\mathsf T}r,\qquad r(0)=-J_Da.
\label{eq:selection_gradient_flow}
\end{equation}
The initial residual lies in $\operatorname{range}L$, where $LL^{\mathsf T}$ is positive definite. It therefore decays exponentially, and $z$ converges to an interpolant. Since all updates lie in $\operatorname{range}L^{\mathsf T}$, the limit is orthogonal to $\ker L$: it is the unique minimum-Euclidean-norm interpolant. This is the standard minimum-norm property of linear gradient flow; the following decomposition identifies its relational component.

The interpolation condition is $J_D(d+Kw-a)+Rw=0$. Orthogonality and injectivity of $J_D$ imply
\begin{equation}
Rw=0,\qquad d+Kw=a.
\label{eq:selection_equivalent_minima}
\end{equation}
This gives $(d,w)=(a-Kv,v)$, $v\in\ker R$. Let the columns of $V$ be an orthonormal basis of $\ker R$, so $VV^{\mathsf T}=\Pi$ and $w=Vv$. The minimum-norm problem becomes
\[
\min_{d,v}\tfrac12(\|d\|_2^2+\|v\|_2^2)
\quad\text{subject to}\quad d+KVv=a.
\]
The multiplier $\zeta$ gives $d=\zeta$, $v=V^{\mathsf T}K^{\mathsf T}\zeta$, and $(I+K\Pi K^{\mathsf T})\zeta=a$. Consequently,
\begin{equation}
d_*=(I+C)^{-1}a,\qquad
w_*=\Pi K^{\mathsf T}(I+C)^{-1}a,\qquad C=K\Pi K^{\mathsf T}.
\label{eq:selection_full_solution}
\end{equation}

For the structural statements, put $T_C=(I+C)^{-1}$, which is symmetric positive definite. For a coordinate support projector $P_S$, no leakage outside $S$ for any $a$ supported on $S$ is equivalent to $(I-P_S)T_CP_S=0$. Symmetry gives the transpose condition, so $T_C$, and hence $C$, is block diagonal across $S,S^c$. Requiring this for every singleton makes $C$ diagonal. Conversely, diagonal $C$ gives $(d_*)_i=a_i/(1+C_{ii})$, preserving every support exactly. If $T_Ca$ is parallel to $a$ for every $a\ne0$, applying the condition to basis vectors and then to $e_i+e_j$ forces $T_C=\alpha I$. Since $C\succeq0$, this is equivalent to $C=\kappa I$ with $\kappa\ge0$. The converse follows from Eq.~\eqref{eq:selection_full_solution}.
\end{proof}

Relational allocation is non-unique exactly when $K\Pi\ne0$; every $d\in\mathbb R^m$ admits compensation exactly when $\operatorname{rank}(K\Pi)=m$. For example, $J_D=I$ and $J_\theta=\sqrt\kappa I$, $\kappa>0$, allow any $d$ while selecting $d_*=a/(1+\kappa)$. Non-identifiability therefore coexists with exact preservation of the target geometry. It does not by itself force such preservation: for $a=(1,0)^{\mathsf T}$ and
$C=\left(\begin{smallmatrix}2&1\\1&2\end{smallmatrix}\right)$,
Eq.~\eqref{eq:relational_selection} gives $d_*=(3/8,-1/8)^{\mathsf T}$.

\subsection{Nearly isotropic compensation and strict ranking}
\label{app:selection_ranking}
\begin{corollary}[Perturbed geometry]
\label{cor:selection_ranking}
Let $C=\kappa I+E_C$, with $\kappa\ge0$ and $\|E_C\|_2\le\varepsilon<1+\kappa$. Define $\alpha=(1+\kappa)^{-1}$ and $\beta=\varepsilon/(1+\kappa-\varepsilon)$. Then
\begin{equation}
\|d_*-\alpha a\|_2\le\alpha\beta\|a\|_2.
\label{eq:selection_ranking_bound}
\end{equation}
If $S=\operatorname{supp}(a)$ is nonempty and proper and $\min_{i\in S}|a_i|>2\beta\|a\|_2$, every entry of $|d_*|$ on $S$ exceeds every entry on $S^c$.
\end{corollary}
\begin{proof}
For $q=1+\kappa$, the resolvent identity gives $(qI+E_C)^{-1}-q^{-1}I=-q^{-1}(qI+E_C)^{-1}E_C$, of norm at most $\varepsilon/[q(q-\varepsilon)]$. This proves the first inequality. Every coordinate error is at most $\alpha\beta\|a\|_2$, so on-support magnitudes are at least $\alpha\min_{i\in S}|a_i|-\alpha\beta\|a\|_2$ and off-support magnitudes are at most $\alpha\beta\|a\|_2$.
\end{proof}

Weak compensation gives a separate sufficient condition without isotropy: since $C\succeq0$,
\begin{equation}
\|d_*-a\|_2
\le\frac{\|C\|_2}{1+\|C\|_2}\|a\|_2.
\label{eq:weak_compensation}
\end{equation}
This follows from $d_*-a=-(I+C)^{-1}Ca$ and the eigenvalues $c_i/(1+c_i)$ of $(I+C)^{-1}C$. Neither result imposes a sparse target; for dense $a$, the same bounds control coefficient distortion.

\subsection{General targets, initialization, and optimization geometry}
\label{app:selection_scope}
For a realizable target $y\in\operatorname{range}[J_D\ J_\theta]$, not necessarily in $\operatorname{range}J_D$, set $a=J_D^\dagger y$, $y_\perp=(I-P_D)y$, and $w_\perp=R^\dagger y_\perp$. The constraints become $w=w_\perp+v$, $v\in\ker R$, and $d+Kv=a-Kw_\perp$. Since $w_\perp\perp\ker R$, the same norm minimization yields
\begin{equation}
d_*=(I+C)^{-1}(a-Kw_\perp),\qquad
w_*=w_\perp+\Pi K^{\mathsf T}d_*.
\label{eq:selection_general_target}
\end{equation}
Thus fitting output components outside the relational tangent subspace can shift the selected relational target.

For arbitrary initialization $z_0$ and realizable target $y$, linear Euclidean gradient flow instead selects
\begin{equation}
z_*=z_0+L^\dagger(y-Lz_0).
\label{eq:selection_initialization}
\end{equation}
Indeed $z-z_0$ remains in $\operatorname{range}L^{\mathsf T}$, and the residual converges to zero as in Eq.~\eqref{eq:selection_gradient_flow}. The initial null-space component is retained. If $d,w$ are displacements around a nonzero network reference, the selection formula describes those displacements; the reference must be added before interpreting final matrix entries.

For a fixed positive-definite preconditioner $P$, the substitution $z=P^{1/2}v$ shows that zero-initialized flow selects $PL^{\mathsf T}(LPL^{\mathsf T})^\dagger y$, minimizing $z^{\mathsf T}P^{-1}z$. More generally, for a differentiable output loss, the Euclidean minimum-distance identity still holds if its flow converges to an exact interpolant, because updates remain in $\operatorname{range}L^{\mathsf T}$. This includes convergent, interpolating Smooth $L_1$ flow in the frozen model. The main result characterizes an unregularized frozen model, rather than the changing Jacobians, adaptive metric, or weight decay of full-network AdamW training.

\subsection{GLV specialization and alignment bounds}
\label{app:glv_alignment_proof}
After separating the local GLV terms, let $a=\operatorname{vec}(A_{\rm off})$ and $d=\operatorname{vec}(D^{\mathsf T}_{\rm off})$ in the same target--source order. The analysis uses the frozen prediction model $\widehat y=\Phi(cd+Kw)$ and $y=\Phi a+\epsilon$, with $c\ne0$ and $\Phi^{\mathsf T}\Phi\succ0$. These are coordinates for analyzing prediction, not additional coefficient or derivative labels used in training.

\begin{proof}[Proof of Corollary~\ref{thm:implicit_D_alignment}]
Decompose $\epsilon=\Phi\delta a+\epsilon_\perp$, where $\delta a=\Phi^\dagger\epsilon$ and $\epsilon_\perp\perp\operatorname{range}\Phi$. Then
\[
\|\Phi(cd+Kw)-y\|_2^2
=\|\Phi(cd+Kw-a-\delta a)\|_2^2+\|\epsilon_\perp\|_2^2.
\]
The second term has zero parameter gradient. Apply Proposition~\ref{prop:relational_selection} with $J_D=c\Phi$, $J_\theta=\Phi K$, and target coordinate $(a+\delta a)/c$. Here $R=0$ and the general compensation matrix is $KK^{\mathsf T}/c^2$. Equation~\eqref{eq:local_allocation} follows. In the noiseless case, $d_*=\alpha a$ with $a\ne0$ is equivalent to $Ba=(c/\alpha-c^2)a$. Thus alignment holds precisely when $a$ is an eigenvector of $B$, and the stated scale follows.
\end{proof}

For a quantitative bound, define
\begin{equation}
\kappa_a=\frac{a^{\mathsf T}Ba}{\|a\|_2^2},\qquad
r_a=Ba-\kappa_a a,\qquad
\alpha=\frac{c}{c^2+\kappa_a},\qquad
\mu=c^2+\lambda_{\min}(B).
\label{eq:glv_alignment_definitions}
\end{equation}
The residual $r_a$ is orthogonal to $a$. Writing $M_c=c^2I+B$, we have $M_c(\alpha a)=ca+\alpha r_a$, hence
\begin{equation}
d_*-\alpha a=M_c^{-1}(c\delta a-\alpha r_a),\qquad
\|d_*-\alpha a\|_2\le |\alpha|\mathcal E,
\label{eq:local_alignment_bound}
\end{equation}
where
\begin{equation}
\mathcal E=
\frac{\|r_a\|_2+(c^2+\kappa_a)\|\delta a\|_2}{\mu}.
\label{eq:glv_alignment_error}
\end{equation}
The bound follows from $M_c\succeq\mu I$ and $|c|/|\alpha|=c^2+\kappa_a$. The controlling quantity is the part of $Ba$ that rotates the target direction, rather than the total strength of compensation.

\paragraph{Pearson correlation and signed weights.} Let $H_m=I-\mathbf1\mathbf1^{\mathsf T}/m$. When $H_ma\ne0$ and $\mathcal E<\|H_ma\|_2$,
\begin{equation}
|\operatorname{corr}(d_*,a)|
\ge\sqrt{1-\left(\frac{\mathcal E}{\|H_ma\|_2}\right)^2}.
\label{eq:glv_pearson_bound}
\end{equation}
To see this, put $u=H_ma/\|H_ma\|_2$ and $v=H_md_*/(\alpha\|H_ma\|_2)$. Orthogonal projection cannot increase norm, so $\|v-u\|_2\le\mathcal E/\|H_ma\|_2<1$. Consequently $v\ne0$ and $u^{\mathsf T}v>0$; the distance from $u$ to the line spanned by $v$ is at most this error. The corresponding angle has sine at most $\mathcal E/\|H_ma\|_2$, proving the bound, including a possible global sign reversal from $\alpha$.

Coordinate-wise, $|(d_*)_i-\alpha a_i|\le|\alpha|\mathcal E$. Entries with $|a_i|>\mathcal E$ retain their sign relative to $\operatorname{sign}(c)$. If $a$ has a nonempty proper support $S$ and $\min_{i\in S}|a_i|>2\mathcal E$, every on-support magnitude exceeds every off-support magnitude. Weak coefficients may therefore have large relative deviations while strong coefficients remain aligned. Applying this coefficient result to Lorenz--96 would require a separate compatible-coordinate argument; its observed support recovery is not obtained by relabelling $a$ as the true graph.

\paragraph{Finite-time observations.} Write the GLV law as $f_i(x)=\ell_i(x)+\sum_{j\ne i}a_{ij}x_ix_j$, with local terms $\ell_i$. On a common trajectory domain, assume $\|Df(x)f(x)\|_\infty\le M_2$. Twice integrating the flow derivative gives
\begin{equation}
\varphi_\Delta(x)-x
=\Delta f(x)+\int_0^\Delta(\Delta-s)
 Df(\varphi_s(x))f(\varphi_s(x))\,\mathrm ds.
\label{eq:glv_finite_time_remainder}
\end{equation}
Stack $n$ observed scalar increments, subtract their local terms $\Delta\ell_i$, and normalize rows by $n^{-1/2}$. If $Z$ contains the correspondingly normalized $x_ix_j$ features in their target blocks, then $y=\Delta Za+\epsilon_\Delta$ with $\|\epsilon_\Delta\|_2\le M_2\Delta^2/2$. For $\sigma_{\min}(Z)=s_0>0$, $\Phi=\Delta Z$, and additional normalized output discrepancy $\epsilon_{\rm other}$,
\begin{equation}
\|\delta a\|_2
\le\frac{M_2\Delta}{2s_0}
  +\frac{\|\epsilon_{\rm other}\|_2}{\Delta s_0}.
\label{eq:glv_observation_error}
\end{equation}
This gives a local analysis model for endpoint data. Errors in the local terms or coefficient parameterization enter $\epsilon_{\rm other}$; the implemented model is still trained on trajectory predictions.

\subsection{Connection to zero-anchored responses}
\label{app:selection_centering}
At a fixed state, the centered branch $C_\theta(q,D)=R_\theta(q,\rho_\theta(q,D))-R_\theta(q,0)$ satisfies $C_\theta(q,0)=\partial_\theta C_\theta(q,0)=0$. Bounded mixed derivatives then give $\|\partial_\theta C_\theta(q,D)\|=O(\|D\|)$ near zero, whereas $\partial_D C_\theta(q,0)$ may be nonzero. This motivates examining unequal relational and response sensitivities, but does not imply the compensation geometry of Proposition~\ref{prop:relational_selection}.

\section{Propagation and Persistent Responses}
\label{app:shared_relations}
We prove the fitted-field propagation and response results, retain the truncation and reset bounds as supplementary lemmas, and connect these properties to the Lorenz--96 diagnostics.

\subsection{Common setup: model, domain, and norms}
\label{app:relational_setup}
For $N$ nodes, let $q_j=(r_j,\eta_j)$, with $r_j=(z_j,u_j)$, and write
\begin{align}
\rho_j(Q)&=\sum_{i\ne j}D_{ij}\Psi_i(q_i),
\label{eq:app_aggregation}\\
\dot r_j&=f_j(q_j,b_j,\rho_j),\qquad
\dot\eta_j=\Lambda_j\{G_j(q_j,b_j,\rho_j)-\eta_j\}.
\label{eq:app_augmented_field}
\end{align}
$D_{ij}$ denotes the directed relation from source $i$ to target $j$, and $D_{jj}=0$. The implemented source map is shared across nodes; the propagation identity also permits node-specific $\Psi_i$. For AugRelNet it has the form $\Psi(q)=(s(q),s(q)m(q))$, where $m$ is a node-local normalized message. The state $u$ participates in the source and receiver maps, and evolves jointly with $z$. The response state $\eta$ participates in those maps and has the additional response equation in Eq.~\eqref{eq:app_augmented_field}. In the Lorenz--96 implementation,
\begin{equation}
G_j(q_j,b_j,\rho_j)
=R_j(q_j,b_j,\rho_j)-R_j(q_j,b_j,0).
\label{eq:app_centered_response}
\end{equation}
Thus $G_j(q_j,b_j,0)=0$. Each $\Lambda_j$ is a fixed positive diagonal matrix. The conditioning vector $b_j$ is fixed during the comparison and enters only node $j$'s receiver. Every cross-node state dependence passes through Eq.~\eqref{eq:app_aggregation}.

Let $\mathcal V_D$ denote the full augmented vector field and $\Phi_D^t$ its flow. We assume all vector fields and source maps are $C^2$ on an open neighborhood of a common compact set $\mathcal Q$. For each comparison below, $\mathcal Q$ contains the relevant trajectories, line segments used in the Lipschitz estimates, and Heun predictor and corrector states. All states lie strictly inside the implementation's clamp range. Derivatives and constants are uniform on this neighborhood. These are finite-horizon assumptions; no global stability or attraction assumption is made.

We use Euclidean norms within a node and the block-maximum norm $\|Q\|_{\mathrm{blk}}=\max_j\|q_j\|_2$ across nodes. Operator norms are induced by the indicated state norms. Write $K\geq0$ for a common Lipschitz bound for the compared full vector fields, and $L_\rho$ for a uniform bound on $\|\partial_{\rho_j}V_j\|$. Constants may depend on the compact domain, parameters, and finite time interval, but not on the initial state within the specified set. Bounds can be taken uniformly over a compact set of fixed conditioners $b$.

\subsection{Directed relational propagation: proof of Proposition~\ref{prop:directed_relational_propagation}}
\label{app:directed_propagation_proof}
Along a trajectory of the actual matrix $D$, the off-diagonal Jacobian blocks factor as
\begin{equation}
\begin{aligned}
\partial_{q_i}\mathcal V_{D,j}(Q(t))&=D_{ij}B_{ji}(t),\quad i\ne j,\\
B_{ji}(t)&=\partial_{\rho_j}V_j\,\partial_{q_i}\Psi_i .
\end{aligned}
\label{eq:app_jacobian_factor}
\end{equation}
The linear aggregation does not force the physical field to be linear or pairwise additive. For two distinct sources $i,k\ne j$, and directions $a_i,a_k$ in their respective augmented states,
\begin{equation}
\begin{split}
\partial_{q_i}\partial_{q_k}V_j[a_i,a_k]
={}&D_{ij}D_{kj}\,\partial_{\rho_j}^2V_j\\
&\left[\partial_{q_i}\Psi_i\,a_i,
      \partial_{q_k}\Psi_k\,a_k\right].
\end{split}
\label{eq:app_mixed_response}
\end{equation}
This follows by applying the chain rule twice to Eq.~\eqref{eq:app_aggregation}; distinct source maps have no mixed state derivative. Thus nonlinear interactions between sources can arise in a shared receiver even though each source is routed by one matrix entry. The receiver, rather than a separate parameter for each interaction term, forms these cross-source products.

Let $A_j(t)$ be the derivative of $V_j$ with respect to its local state, holding $\rho_j$ fixed. Because $D_{jj}=0$, it is the diagonal block of the full Jacobian. The variational equation is
\begin{equation}
\dot\xi_j(t)=A_j(t)\xi_j(t)
+\sum_{i\ne j}D_{ij}B_{ji}(t)\xi_i(t).
\label{eq:app_variation}
\end{equation}
The matrices $A_j$ include all local $z,u,\eta$ couplings. Let $U_j(t,s)$ solve $\partial_tU_j(t,s)=A_j(t)U_j(t,s)$ with $U_j(s,s)=I$. Boundedness gives constants $L,M\geq0$ such that
\begin{equation}
\|U_j(t,s)\|\leq e^{L(t-s)},\qquad
\|B_{ji}(t)\|\leq M,
\label{eq:app_local_bounds}
\end{equation}
for $0\le s\le t\le T$.

Let $\mathcal W^r_{i\to j}$ be the set of directed length-$r$ walks $p=(v_0=i,v_1,\ldots,v_r=j)$ through nonzero entries of $D$. Nodes may repeat. Define $D_p=\prod_{\ell=1}^rD_{v_{\ell-1}v_\ell}$. Put $s_0=0$, $\Delta_r(t)=\{0<s_1<\cdots<s_r<t\}$, and
\[
C_\ell=B_{v_\ell v_{\ell-1}}(s_\ell)
U_{v_{\ell-1}}(s_\ell,s_{\ell-1}).
\]
The time-ordered kernel is
\begin{equation}
\mathcal K_p(t)=
\int_{\Delta_r(t)}
U_{v_r}(t,s_r)
\prod_{\ell=r}^{1}C_\ell
\,\mathrm ds_1\cdots\mathrm ds_r,
\label{eq:app_walk_kernel}
\end{equation}
where the product is ordered from $\ell=r$ on the left to $\ell=1$ on the right. Each factor has the appropriate node-block dimensions. The propagation intervals in the integrand sum to $t$. The ordered simplex has volume $t^r/r!$, so
\begin{equation}
\|\mathcal K_p(t)\|\leq e^{Lt}M^rt^r/r!.
\label{eq:app_kernel_bound}
\end{equation}

Variation of constants in Eq.~\eqref{eq:app_variation}, followed by successive substitution of the integral term, gives
\begin{equation}
T^D_{ji}(t)=\mathbf1_{i=j}U_j(t,0)
+\sum_{r=1}^{\infty}\sum_{p\in\mathcal W^r_{i\to j}}
D_p\mathcal K_p(t).
\label{eq:app_walk_identity}
\end{equation}
Put $W=|D|^\top$, so $W_{ji}=|D_{ij}|$. The absolute sum at order $r$ is at most $e^{Lt}(Mt)^r(W^r)_{ji}/r!$. Its sum is bounded by the corresponding entry of $e^{Lt}e^{MtW}$; hence the series converges absolutely and uniformly on bounded time intervals. The integral remainder tends to zero by the same bound, proving Eq.~\eqref{eq:app_walk_identity}.

For distinct nodes with no directed path, every term is absent and $T^D_{ji}(t)=0$. If the shortest path has length $r$, the first possible term is of order $t^r$, proving the $O(t^r)$ statement. This is an upper-order statement: vanishing derivatives or cancellation can remove the leading term. The kernels are evaluated on the $D$-dependent trajectory and are themselves generally $D$-dependent. Equation~\eqref{eq:app_walk_identity} is therefore a factorization of responses, not a polynomial in $D$ with fixed coefficients.

\subsubsection{A functional truncation bound}
\label{app:functional_tail_proof}
The same routing representation gives a perturbation bound when weak routes are discarded. This bound concerns the continuous realization and its Heun integrator.

\begin{lemma}[Functional tail truncation]
\label{prop:functional_tail_truncation}
Under the setup of Appendix~\ref{app:relational_setup}, retain a directed support $S$ by setting $D^S_{ij}=D_{ij}\mathbf1_{(i,j)\in S}$ and define
\begin{equation}
\tau_S=\max_j\sum_{i:(i,j)\notin S}
|D_{ij}|\,\sup_{Q\in\mathcal Q}\|\Psi_i(q_i)\|_2.
\label{eq:functional_tail}
\end{equation}
Starting from the same full augmented state and fixed conditioner, the flows and $h$ complete Heun steps of size $\delta$ satisfy
\begin{equation}
\begin{aligned}
\|\Phi_D^t-\Phi_{D^S}^t\|_{\mathrm{blk}}
&\le L_\rho\tau_S\frac{e^{Kt}-1}{K},\\
\|Q_h^D-Q_h^{D^S}\|_{\mathrm{blk}}
&\le \frac{L_\rho\tau_S}{K}(r_\delta^h-1)
\le L_\rho\tau_S\frac{e^{Kh\delta}-1}{K},
\end{aligned}
\label{eq:functional_tail_bounds}
\end{equation}
where $r_\delta=1+K\delta+K^2\delta^2/2$. At $K=0$, the respective bounds are $L_\rho\tau_St$ and $h\delta L_\rho\tau_S$.
\end{lemma}

\begin{proof}
The diagonal contributes zero. This is a \emph{functional} tail: it includes the magnitude of emitted source features. In particular it is invariant to the rescaling $D\mapsto cD$, $\Psi_i\mapsto\Psi_i/c$, $c\ne0$. Small raw entries alone are not a scale-invariant compression certificate.

At a common state, the aggregation changes by at most $\tau_S$ at each target. The receiver Lipschitz bound therefore gives
\begin{equation}
\|\mathcal V_D(Q)-\mathcal V_{D^S}(Q)\|_{\mathrm{blk}}
\leq L_\rho\tau_S.
\label{eq:app_field_tail}
\end{equation}
Let the two flows start from the same full augmented state and use the same fixed conditioner. Adding and subtracting $\mathcal V_D(\Phi_{D^S}^t)$ yields
\begin{equation}
e(t)\leq\int_0^t\{K e(s)+L_\rho\tau_S\}\,ds,
\quad
e(t)=\|\Phi_D^t-\Phi_{D^S}^t\|_{\mathrm{blk}}.
\end{equation}
Gronwall's inequality gives
\begin{equation}
\|\Phi_D^t-\Phi_{D^S}^t\|_{\mathrm{blk}}
\leq L_\rho\tau_S\frac{e^{Kt}-1}{K},
\label{eq:app_flow_tail}
\end{equation}
with value $L_\rho\tau_St$ when $K=0$. Both flows retain their complete $u,\eta$ dynamics. Minimizing $\tau_S$ over supports with at most $s$ incoming edges per target defines a best-$s$ functional tail. A small best-$s$ tail certifies closeness to an $s$-neighbor realization without having prescribed that support during training.

\paragraph{The implemented Heun map.} The same bound holds directly for Heun integration; no continuous-to- discrete identification is needed. For a vector field $V$, write
\begin{equation}
\mathcal H_\delta^V(Q)
=Q+\frac{\delta}{2}\{V(Q)+V(Q+\delta V(Q))\}.
\label{eq:app_heun}
\end{equation}
If $V$ is $K$-Lipschitz, this map is at most $r_\delta=1+K\delta+K^2\delta^2/2$ Lipschitz. If $\|V(Q)-\widetilde V(Q)\|\leq c$ uniformly and both fields are $K$-Lipschitz, a same-input step differs by at most $d_\delta=c\delta(1+K\delta/2)$. Indeed the corrector difference is bounded by $c+K\delta c$, while the predictor derivative difference is at most $c$. Iteration from a common initial state gives
\begin{equation}
\begin{aligned}
\|Q_h^D-Q_h^{D^S}\|_{\mathrm{blk}}
&\leq d_\delta\frac{r_\delta^h-1}{r_\delta-1}\\
&=\frac{L_\rho\tau_S}{K}(r_\delta^h-1)\\
&\leq L_\rho\tau_S\frac{e^{Kh\delta}-1}{K}.
\end{aligned}
\label{eq:app_heun_tail}
\end{equation}
Here $c=L_\rho\tau_S$; the $K=0$ limit follows directly from the recursion and equals $h\delta L_\rho\tau_S$.
\end{proof}

\subsection{Relaxation-weighted response approximation}
\label{app:response_approximation}
Fix $D$, $b$, and positive diagonal $\Lambda$, and write
\begin{equation}
\dot X=F_D(X,E),\qquad
\dot E=\Lambda\{G_D(X,E)-E\},\qquad E=\eta.
\label{eq:app_response_system}
\end{equation}
All source-emission and receiver feedback is included in the stacked maps. Use coordinate maximum norms and $\|v\|_{\infty,T}=\sup_{0\le t\le T}\|v(t)\|_\infty$ in this subsection. On a common domain containing the compared trajectories, take nonnegative constants satisfying
\begin{equation}
\begin{aligned}
\|F_D(X,E)-F_D(X',E')\|_\infty
&\le a_x\|X-X'\|_\infty+a_e\|E-E'\|_\infty,\\
\|G_D(X,E)-G_D(X',E')\|_\infty
&\le b_x\|X-X'\|_\infty+b_e\|E-E'\|_\infty.
\end{aligned}
\label{eq:response_lipschitz}
\end{equation}
For $\lambda_{\max}=\max_k\lambda_k$, a valid constant in Proposition~\ref{prop:response_approximation} is
\begin{equation}
M=\begin{pmatrix}
a_x&a_e\\ \lambda_{\max}b_x&\lambda_{\max}b_e
\end{pmatrix},\qquad C_T=(e^{MT})_{12}.
\label{eq:response_comparison_constant}
\end{equation}
The perturbation $e$ may be any bounded measurable drive error whose perturbed trajectory remains in this domain.

\begin{proof}[Proof of Proposition~\ref{prop:response_approximation}]
Let $\Delta X=\widetilde X-X$ and $\Delta E=\widetilde E-E$. Variation of constants, with the same full initial state, gives
\begin{equation}
\Delta E=\mathcal T_\Lambda(\Delta G_D)+\mathcal T_\Lambda e,\qquad
\Delta G_D(t)=G_D(\widetilde X(t),\widetilde E(t))-G_D(X(t),E(t)).
\label{eq:closed_loop_filter_identity}
\end{equation}
This retains the full nonlinear feedback through the perturbed trajectory. Set $x(t)=\|\Delta X(t)\|_\infty$, $z(t)=\|\Delta E(t)\|_\infty$, and $v_0=\|\mathcal T_\Lambda e\|_{\infty,T}$. Equation~\eqref{eq:response_lipschitz} yields
\begin{align}
x(t)&\le\int_0^t\{a_x x(s)+a_e z(s)\}\,\mathrm ds,\\
z(t)&\le v_0+\int_0^t\lambda_{\max}
                   \{b_x x(s)+b_e z(s)\}\,\mathrm ds.
\label{eq:response_comparison_integrals}
\end{align}
The second inequality uses $\|e^{-\Lambda(t-s)}\Lambda\|_\infty\le\lambda_{\max}$, while keeping $\mathcal T_\Lambda e$ intact. For $u=(x,z)^{\mathsf T}$ and $q=(0,v_0)^{\mathsf T}$, these inequalities are $u(t)\le q+\int_0^tMu(s)\,\mathrm ds$ componentwise. Iterating gives $u(t)\le e^{Mt}q$: the remainder tends to zero on $[0,T]$ by the factorial bound for iterated integrals. Since $M$ has nonnegative entries, $(e^{Mt})_{12}$ is nondecreasing, proving Eq.~\eqref{eq:filtered_forecast_error} uniformly on $[0,T]$. An approximate response map $\widetilde G_D$ is covered by setting $e(t)=(\widetilde G_D-G_D)(\widetilde X(t),\widetilde E(t))$.

For the parameter statement, suppose the maps are $C^1$ and perturb only drive parameters $\theta$, keeping $F_D,D,b,\Lambda$ and the initial state fixed. In direction $v$, the state sensitivities $(\xi,\zeta)$ obey
\begin{equation}
\begin{aligned}
\dot\xi&=F_X(t)\xi+F_E(t)\zeta,\\
\dot\zeta&=\Lambda\{G_X(t)\xi+G_E(t)\zeta+\psi_v(t)-\zeta\},
\qquad \xi(0)=\zeta(0)=0,
\end{aligned}
\label{eq:response_parameter_tangent}
\end{equation}
with $\psi_v(t)=\partial_\theta G_D(X(t),E(t))v$. The same comparison proves $\|\xi\|_{\infty,T}\le C_T\|\mathcal T_\Lambda\psi_v\|_{\infty,T}$. For $n_z$ observed coordinates selected by $P_z$, define the normalized multi-horizon forecast
\begin{equation}
\mathcal Y(\theta)=\operatorname{col}_{h=1}^H
 \left(\sqrt{\frac{w_h}{n_z}}P_zX_\theta(t_h)\right),\qquad
w_h\ge0,\quad\sum_hw_h=1.
\label{eq:normalized_prediction_vector}
\end{equation}
Thus $\|J_\eta v\|_2^2=\sum_h(w_h/n_z)\|P_z\xi(t_h)\|_2^2 \le\|\xi\|_{\infty,T}^2$, proving the sensitivity claim.
\end{proof}

\paragraph{Consequences for response fitting.} For fixed forecast targets, the triangle inequality gives
\begin{equation}
\|\widetilde{\mathcal Y}-\mathcal Y^{\rm target}\|_2
\le\|\mathcal Y-\mathcal Y^{\rm target}\|_2
   +C_T\|\mathcal T_\Lambda e\|_{\infty,T}.
\label{eq:response_prediction_risk}
\end{equation}
A good reference predictor therefore tolerates a small \emph{filtered} response error, even when its instantaneous amplitude is larger. Likewise the directional Gauss--Newton quantity satisfies
\begin{equation}
v^{\mathsf T}J_\eta^{\mathsf T}J_\eta v
\le C_T^2\|\mathcal T_\Lambda\psi_v\|_{\infty,T}^2.
\label{eq:response_gauss_newton}
\end{equation}
These bounds characterize response approximation and prediction sensitivity; errors entering $F_D$ directly are not filtered in the same way. The response state supplies this temporal filter explicitly.

\paragraph{Exact tangent factorization.} The integrated form of Eq.~\eqref{eq:response_parameter_tangent} has a bounded Volterra feedback operator $\mathcal V$. Its iterates have factorially decreasing norms on a finite interval, so its resolvent exists and
\begin{equation}
\begin{pmatrix}\xi\\\zeta\end{pmatrix}
=(I-\mathcal V)^{-1}
 \begin{pmatrix}0\\\mathcal T_\Lambda\psi_v\end{pmatrix}.
\label{eq:response_tangent_factorization}
\end{equation}
Thus parameter-to-prediction sensitivity factors through the relaxation filter before the complete feedback resolvent and readout.

\subsubsection{Time scales and cancelling approximation errors}
\label{app:response_timescales}
Because each diagonal kernel integrates to at most one, $\|\mathcal T_\Lambda e\|_{\infty,T}\le\|e\|_{\infty,T}$. For $e(t)=\operatorname{Re}(v e^{i\omega t})$, direct integration gives
\begin{equation}
\mathcal T_\Lambda(v e^{i\omega\cdot})(t)
=\operatorname{diag}\!\left(
\frac{\lambda_k}{\lambda_k+i\omega}
(e^{i\omega t}-e^{-\lambda_kt})\right)v.
\label{eq:response_harmonic_filter}
\end{equation}
Hence, including the initial transient,
\begin{equation}
\|\mathcal T_\Lambda e\|_{\infty,T}
\le\frac{2\lambda_{\max}}{\sqrt{\lambda_{\max}^2+\omega^2}}
\|v\|_\infty.
\label{eq:response_frequency_bound}
\end{equation}
Together with Proposition~\ref{prop:response_approximation}, this controls how rapidly cancelling drive errors affect the full closed-loop forecast. For a nonsinusoidal scalar error, if $E_0(t)=\int_0^t e(s)\,\mathrm ds$ satisfies $\|E_0\|_{\infty,T}\le\varepsilon_0$, integration by parts gives $\|\mathcal T_\lambda e\|_{\infty,T}\le2\lambda\varepsilon_0$.

For a scalar rate, a slowly varying signal $s$ with $\|\dot s\|_{\infty,T}\le\nu$ and initialization $\eta(0)=s(0)$ obeys
\[
\eta(t)-s(t)=-\int_0^t e^{-\lambda(t-r)}\dot s(r)\,\mathrm dr,
\qquad \|\eta-s\|_{\infty,T}\le\nu/\lambda.
\]
An additional harmonic drive error of amplitude $A_e$ therefore yields
\begin{equation}
\|\eta-s\|_{\infty,T}
\le\frac{\nu}{\lambda}
 +\frac{2\lambda A_e}{\sqrt{\lambda^2+\omega^2}}.
\label{eq:response_timescale_tradeoff}
\end{equation}
The rate trades tracking lag against rejection of faster errors. This interpretation requires a separation between useful response variation and approximation error; high-frequency physical dynamics are not intrinsically undesirable.

\subsubsection{The implemented Heun response}
\label{app:response_heun}
The actual two-stage update is
\begin{equation}
\begin{aligned}
k_1&=V_D(q_n,b),& k_2&=V_D(q_n+\delta k_1,b),\\
q_{n+1}&=q_n+\tfrac\delta2(k_1+k_2).
\end{aligned}
\label{eq:method_heun}
\end{equation}
For a scalar rate, put $r=\delta\lambda$, and let $g_n,g_n^p$ be the drive values at the two stages. Substituting the predictor $\eta_n^p=(1-r)\eta_n+rg_n$ into the corrector gives exactly
\begin{equation}
\eta_{n+1}=(1-r+r^2/2)\eta_n
 +\tfrac r2\{(1-r)g_n+g_n^p\}.
\label{eq:heun_response_exact}
\end{equation}
The homogeneous decay factor is below one for $0<r<2$. For $0<r\le1$, the two drive coefficients are nonnegative; when both stage drives coincide, the update is $\eta_{n+1}=q_r\eta_n+(1-q_r)g_n$, $q_r=1-r+r^2/2$. Endogenous stage drives still carry feedback. A positive rate alone therefore does not guarantee stability of the full discrete update, and the continuous bound is not asserted for arbitrary Heun step sizes. The reset lemma below concerns the implemented two-stage intervention separately.

\subsection{The completed-step reset intervention}
\label{app:persistent_response_proof}
Equation~\eqref{eq:eta_memory} remains exact when the drive depends on the response state itself. Positive relaxation rates alone do not ensure closed-loop stability: $G=k\eta$ with $k>1$ gives growth. The following result concerns the specific intervention used in the experiments.

\begin{lemma}[Persistent response under reset]
\label{prop:persistent_response_reset}
Under Appendix~\ref{app:relational_setup}, write $X=(z,u)$, $E=\eta$, $\dot X=F_D(X,E)$ and $\dot E=H_D(X,E)=\Lambda(G_D(X,E)-E)$. Let $X_h$ denote $h$ complete Heun steps of size $\delta$ from $(X_0,E_0)$, and $\bar X_h$ the same computation resetting only $E$ to $E_0$ after each completed step. For fixed $h$ as $\delta\to0$,
\begin{equation}
\begin{aligned}
P(X_h-\bar X_h)&=\tfrac12h(h-1)\delta^2\mathcal B_D
+O_h(\delta^3),\\
\mathcal B_D&=\left.P\,\partial_EF_D\,H_D\right|_{(X_0,E_0)},
\end{aligned}
\label{eq:persistent_response}
\end{equation}
where $P$ selects the observed coordinates. The first completed-step predictions coincide exactly. A nonzero leading coefficient separates these responses for $h\ge2$.
\end{lemma}

\begin{proof}
Stack the states as $\dot X=F_D(X,E)$, $\dot E=H_D(X,E)=\Lambda(G_D(X,E)-E)$. All quantities with subscript zero below are evaluated at $(X_0,E_0)$. We compare the normal Heun recursion with a recursion that completes the same Heun step and then replaces only its output $E$ by $E_0$. In particular the reset recursion still uses the predicted $E+\delta H_D$ inside the Heun corrector.

For a fixed number $h$ of steps and $\delta\to0$, Taylor expansion of one complete step gives
\begin{equation}
X^+=X+\delta F_D
+\frac{\delta^2}{2}\left(F_XF_D+F_EH_D\right)
+O(\delta^3).
\label{eq:app_heun_X}
\end{equation}
The uniform $O(\delta^3)$ remainder follows from the $C^2$ assumption on the compact stage domain. Inductively, for $0\leq n\leq h$, the full recursion satisfies $X_n=X_0+n\delta F_0+O_h(\delta^2)$ and $E_n=E_0+n\delta H_0+O_h(\delta^2)$. Substitution into Eq.~\eqref{eq:app_heun_X} shows that the coefficient of $\delta^2(F_XF_0+F_EH_0)$ at step $n$ is $n+1/2$. Since $\sum_{n=0}^{h-1}(n+1/2)=h^2/2$,
\begin{equation}
X_h=X_0+h\delta F_0
+\frac{h^2\delta^2}{2}(F_XF_0+F_EH_0)
+O_h(\delta^3).
\label{eq:app_full_heun_expansion}
\end{equation}

The reset recursion starts every step at $E_0$, and $\bar X_n=X_0+n\delta F_0+O_h(\delta^2)$. The first-stage expansion contributes $nF_XF_0$ at order $\delta^2$, whereas the corrector contributes $(F_XF_0+F_EH_0)/2$. Consequently
\begin{equation}
\bar X_h=X_0+h\delta F_0
+\frac{\delta^2}{2}\left(h^2F_XF_0+hF_EH_0\right)
+O_h(\delta^3).
\label{eq:app_reset_heun_expansion}
\end{equation}
Subtracting Eqs.~\eqref{eq:app_full_heun_expansion} and \eqref{eq:app_reset_heun_expansion}, and applying $P$, proves Eq.~\eqref{eq:persistent_response}. At $h=1$ the physical outputs coincide exactly, because reset occurs after the common completed step.

The derivative $F_E$ is the derivative of the \emph{complete} stacked field. It includes both the target-local dependence on $\eta$ and the dependence through source emission and aggregation. The coefficient $\mathcal B_D$ therefore measures how the response update returns to the observed dynamics, including its relational routes. If $\mathcal B_D=0$, this second-order term vanishes; higher-order $\eta$ dependence may still be present. This lemma characterizes the completed-step reset intervention; separately trained state ablations are reported in Table~\ref{tab:l96_state_ablations}.
\end{proof}

\subsubsection{A combined truncation--reset comparison}
\label{app:dynamic_reset_comparison}
The bounds of Lemmas~\ref{prop:functional_tail_truncation} and~\ref{prop:persistent_response_reset} can be combined in the same prediction norm. For a distribution $\mu$ of initial augmented states, define
\begin{equation}
\|Y\|_{2,\mu,N}
=\left(\frac1N\mathbb E_\mu\sum_{j=1}^N|Y_j|^2\right)^{1/2}.
\end{equation}
Let $y^\star_h$ be the forecast target, and let $y_D$, $y_{D^S}$, and $y_{\mathrm{reset}}$ be the normal, support-truncated dynamic, and dense reset Heun predictions. Put
\begin{align}
e_D&=\|y_D-y^\star_h\|_{2,\mu,N},\qquad
a=\|\mathcal B_D\|_{2,\mu,N},\\
T_S&=L_\rho\tau_S(e^{Kh\delta}-1)/K,\qquad
B_h=\tfrac12h(h-1)a\delta^2-C_h\delta^3,
\label{eq:app_comparison_constants}
\end{align}
where $C_h$ bounds the uniform remainder in Eq.~\eqref{eq:persistent_response}. The projection to one observed coordinate per node and normalization by $N$ ensure that the block-maximum bound in Eq.~\eqref{eq:app_heun_tail} also bounds this prediction norm. All errors here use the same observed-coordinate units; a further linear change of units multiplies the bounds by the corresponding operator norm. The triangle and reverse-triangle inequalities give
\begin{equation}
\begin{aligned}
\|y_{D^S}-y^\star_h\|_{2,\mu,N}&\leq e_D+T_S,\\
\|y_{\mathrm{reset}}-y^\star_h\|_{2,\mu,N}&\geq B_h-e_D.
\end{aligned}
\label{eq:app_prediction_comparison}
\end{equation}
Hence $B_h>2e_D+T_S$ is sufficient for the smaller, dynamic realization to be strictly more accurate than the dense reset realization. This condition separates three quantities: the original prediction error, the removable functional tail, and the response discarded by resetting. Accuracy does not follow from sparsity or augmentation alone. The result states when preserving the dynamic realization is more valuable than retaining every weak relation.

\subsection{Why structural zeros are robust to the receiver}
\label{app:structural_zeros}
At a fixed augmented state, a zero $D_{ij}$ makes the instantaneous cross-node derivative in Eq.~\eqref{eq:app_jacobian_factor} zero for every smooth receiver map. A nonzero edge may also have zero derivative, but then the zero can depend on the receiver. The following identity makes this distinction quantitative.

Let $\mu$ be a \emph{fixed} distribution of augmented states supported in $\mathcal Q$. For a scalar observed coordinate $z_j$, perturb the receiver function by
\begin{equation}
\dot z_j^{\varepsilon}=\dot z_j+\varepsilon v^\top\rho_j,
\qquad \mathbb E v=0,\quad\mathbb E vv^\top=I,
\label{eq:app_receiver_perturbation}
\end{equation}
where $v$ is independent of the state. A common $v$ can be used for every target, preserving a shared receiver perturbation. Define the off-support instantaneous derivative energy
\begin{equation}
\mathcal E_{S^c}
=\sum_{i\ne j:(i,j)\notin S}
\mathbb E_\mu\|\partial_{q_i}\dot z_j\|_2^2.
\end{equation}
Then
\begin{equation}
\mathbb E_v\mathcal E_{S^c}^{\varepsilon}
=\mathcal E_{S^c}
+\varepsilon^2\sum_{i\ne j:(i,j)\notin S}
D_{ij}^2\,\mathbb E_\mu\|\partial_{q_i}\Psi_i(q_i)\|_F^2.
\label{eq:app_robust_zero}
\end{equation}
To prove this, differentiate Eq.~\eqref{eq:app_receiver_perturbation}: $\partial_{q_i}\dot z_j^{\varepsilon} =\partial_{q_i}\dot z_j+\varepsilon D_{ij}v^\top\partial_{q_i}\Psi_i$. Expansion of the squared norm gives a cross term whose expectation is zero. Isotropy gives $\mathbb E_v\|v^\top A\|_2^2 =\operatorname{tr}(AA^\top)=\|A\|_F^2$, proving the identity.

If $\mathbb E_\mu\|\partial_{q_i}\Psi_i\|_F^2>0$, a nonzero edge contributes a strictly positive perturbation term even when its original receiver cancels the instantaneous response. A zero edge contributes none. Structural zeros are therefore robust to this receiver perturbation, whereas cancellations at nonzero edges are not. The perturbation changes the receiver while holding $D$ and the state distribution $\mu$ fixed.

\subsection{A link between response filtering and relational selection}
\label{app:filtered_selection}
Use the same normalized continuous-time forecast coordinates in Propositions~\ref{prop:relational_selection} and~\ref{prop:response_approximation}, and split $J_\theta=[J_\eta\ J_o]$ into drive-only and remaining parameter columns. In the compatible frozen model, suppose $\sigma_{\min}(J_D)\ge s_0>0$, $\|J_o\|_2\le b_0$, and $\sup_{\|v\|_2=1}\|\mathcal T_\Lambda\psi_v\|_{\infty,T} \le\delta_\eta$. Then
\begin{equation}
\|C\|_2\le q,\qquad
\|d_*-a\|_2\le\frac{q}{1+q}\|a\|_2,\qquad
q=\frac{b_0^2+C_T^2\delta_\eta^2}{s_0^2}.
\label{eq:filtered_selection_link}
\end{equation}
Indeed, Proposition~\ref{prop:response_approximation} gives $\|J_\eta\|_2\le C_T\delta_\eta$. Thus $\|J_\theta\|_2^2 \le\|J_\eta\|_2^2+\|J_o\|_2^2 \le C_T^2\delta_\eta^2+b_0^2$ and $\|K\Pi K^{\mathsf T}\|_2 \le\|J_D^\dagger\|_2^2\|J_\theta\|_2^2\le q$. Equation~\eqref{eq:weak_compensation} completes the argument. This gives a sufficient weak-compensation regime; changing $\Lambda$ also changes the trajectory, $J_D$, $J_o$, and $C_T$.

\subsection{Lorenz--96 and discrete observation intervals}
\label{app:l96_theory_scope}
The physical Lorenz--96 vector field has off-diagonal derivatives
\begin{equation}
\begin{aligned}
\partial_{x_{j-1}}f_j&=x_{j+1}-x_{j-2},\\
\partial_{x_{j+1}}f_j&=x_{j-1},\qquad
\partial_{x_{j-2}}f_j=-x_{j-1}.
\end{aligned}
\label{eq:app_l96_derivatives}
\end{equation}
and all other cross-variable derivatives are zero. Thus the instantaneous support is fixed while its magnitudes and signs vary with state. In particular the last two derivatives are opposed whenever $x_{j-1}\ne0$. On a common compact flow domain, $\partial_{x_i}\varphi_{t,j} =\mathbf1_{i=j}+t\partial_{x_i}f_j+O(t^2)$ uniformly. For example, if $\mathbb E[(x_{j+1}-x_{j-2})^2]>0$ and $\mathbb E[x_{j-1}^2]>0$, the three coordinate responses have nonzero first-order $L^2$ terms, whereas nonneighbor coordinate responses are $O(t^2)$.

Proposition~\ref{prop:directed_relational_propagation} describes how longer observation intervals allow longer walks to contribute to the measured flow response. This occurs in a continuous model: the discrete object is the set of observation times. For an exactly sparse support, path length determines the first possible order of a response. For a dense matrix with a sparse dominant pattern, the statement applies to its support-truncated realization, with the tail error controlled by Lemma~\ref{prop:functional_tail_truncation} (Eq.~\eqref{eq:app_flow_tail}). Observation spacing therefore changes which finite-time pathways can contribute, without imposing a monotonic edge-addition rule on the fitted $D$.

The proposition characterizes finite-time propagation in the fitted model. Recovered support and relative signs are empirical properties of the trained realization, consistent with persistent relational organization and state-dependent response.

\paragraph{Integration and conditioning scope.} The short-step expansion in Lemma~\ref{prop:persistent_response_reset} uses fixed $h$ and $\delta\to0$. In the reported Lorenz--96 implementation one observation interval $\Delta=0.02$ is represented by a Heun step of size $1$ in learned model time. The asymptotic coefficient is therefore not used to predict the numerical magnitude of the $h=8$ or $h=40$ intervention ratios. Those ratios are measured directly. Equations~\eqref{eq:app_walk_identity} and~\eqref{eq:app_flow_tail} concern the continuous field, while Eq.~\eqref{eq:app_heun_tail} concerns its Heun integrator. The continuous response-error bound in Proposition~\ref{prop:response_approximation} is subject to the discretization scope in Appendix~\ref{app:response_heun}. A rolling conditioner adds feedback beyond the fixed-$b$ comparison.

\section{Local Physical Support and Predictive Dependence in Ising}
\label{app:ising_structure}

The Ising experiment distinguishes microscopic interaction from predictive dependence at the recorded time scale. Two sources of broader dependence must be separated: spatial correlation under the Gibbs distribution and composition of updates within a recorded sweep. The proposition below isolates the former for a single-site update; we then connect the actual full-sweep transition to the forecasting and intervention diagnostics.

Consider a zero-field ferromagnetic Ising model on a finite lattice, with spins $X_v\in\{-1,+1\}$ and Gibbs distribution
\[
\mu_\beta(x)
\propto
\exp\!\left(
    \beta J\sum_{\langle u,v\rangle}x_u x_v
\right),
\qquad J>0.
\]
Let $\mathcal N(i)$ denote the four nearest neighbours of site $i$. Suppose $X\sim\mu_\beta$, and let $Y_i$ be obtained by a heat-bath resampling of site $i$ conditional on the remaining spins.

\begin{proposition}
\label{prop:ising_predictive_dependence}
For a single-site heat-bath update at site $i$:

\begin{enumerate}
\item The microscopic transition is local:
    \[
    \Pr(Y_i=s\mid X)
    =
    \Pr(Y_i=s\mid X_{\mathcal N(i)})
    =
    \frac{
        \exp\!\left(
            \beta J s\sum_{j\in\mathcal N(i)}X_j
        \right)
    }{
        2\cosh\!\left(
            \beta J\sum_{j\in\mathcal N(i)}X_j
        \right)
    }.
    \]
Hence, for every $k\notin\mathcal N(i)\cup\{i\}$,
    \[
    I(Y_i;X_k\mid X_{\mathcal N(i)})=0.
    \]

\item Nevertheless, heat-bath resampling preserves the joint distribution of the updated spin with every unupdated spin:
    \[
    (Y_i,X_k)\overset{d}{=}(X_i,X_k),
    \qquad k\neq i.
    \]
Consequently,
    \[
    I(Y_i;X_k)=I(X_i;X_k).
    \]
\end{enumerate}
\end{proposition}

\begin{proof}
The first statement follows directly from the nearest-neighbour heat-bath conditional distribution, which depends on the current configuration only through $X_{\mathcal N(i)}$.

For the second statement, let $f$ be any bounded function of one spin and the configuration outside site $i$. Since $Y_i$ is sampled from the Gibbs conditional distribution of $X_i$ given $X_{-i}$,
\[
\begin{aligned}
\mathbb E[f(Y_i,X_{-i})]
&=
\mathbb E\!\left[
    \sum_{s\in\{-1,+1\}}
    \mu_\beta(X_i=s\mid X_{-i})
    f(s,X_{-i})
\right] \\
&=
\mathbb E[f(X_i,X_{-i})].
\end{aligned}
\]
Thus
\[
(Y_i,X_{-i})\overset{d}{=}(X_i,X_{-i}),
\]
and marginalizing to any $k\neq i$ gives
\[
(Y_i,X_k)\overset{d}{=}(X_i,X_k).
\]
Mutual information is determined by the joint distribution, yielding $I(Y_i;X_k)=I(X_i;X_k)$.
\end{proof}

For this single-site update, Gibbs correlations allow non-neighbours to carry marginal predictive information even though the four direct neighbours suffice conditionally. This distinguishes microscopic support from marginal dependence without invoking multi-step propagation.

\paragraph{The recorded full-sweep transition.} One experimental step updates both checkerboard sublattices, with their order drawn independently and uniformly for each sweep. Write $Z_i$ for the spin after this complete sweep, to distinguish it from the single-site $Y_i$ above, and define
\[
p_i(x)=\sigma\!\left(2\beta J\sum_{k\in\mathcal N(i)}x_k\right).
\]
If $i$ updates first, its probability of $+1$ is $p_i(x)$. If it updates second, its four neighbours have already been independently resampled conditional on $x$, with probabilities $p_k(x)$. Marginalizing those intermediate spins and the unobserved order gives the exact sitewise oracle
\begin{equation}
\begin{aligned}
q_i(x)&:=\Pr(Z_i=+1\mid X=x)\\
&=\tfrac12 p_i(x)+\tfrac12
\sum_{z\in\{-1,+1\}^{\mathcal N(i)}}
\sigma\!\left(2\beta J\sum_{k\in\mathcal N(i)}z_k\right)
\prod_{k\in\mathcal N(i)}
p_k(x)^{(1+z_k)/2}[1-p_k(x)]^{(1-z_k)/2}.
\end{aligned}
\label{eq:ising_full_sweep_oracle}
\end{equation}
The sum enumerates 16 outcomes and is valid for any observed input, without a stationarity assumption. The second term depends on old spins two lattice edges from $i$, including diagonals and axial distance-two sites, as well as on $x_i$. Thus the old four-neighbour configuration need not suffice for a full recorded step. Although complete sweeps preserve the Gibbs distribution, stationarity of one-time marginals does not imply the mixed-time identity $(Z_i,X_k)\overset d=(X_i,X_k)$ used for the single-site result.

The cardinal neighbours therefore retain their microscopic role, while diagonal inputs can carry both correlated information and composed transition dependence. As in coarse-sampled Lorenz--96, local updates can produce a broader finite-step predictor, here alongside equilibrium correlations. The learned cardinal preference is an empirical distinction within this predictive neighbourhood, rather than a requirement that every diagonal route vanish. The interventions below test its functional value; they do not apportion the two sources of predictive information.

\subsection{Message redundancy and weight interventions}
\label{app:ising_diagnostics}
The following diagnostics complement the headline forecasting and cardinal--diagonal intervention in Section~\ref{sec:ising}.

Weight permutation acts through the aggregated signal,
\[
\rho=\sum_\delta D_\delta\Psi_\delta .
\]
Across five seeds, random permutations change $D$ by $22.5\%$ in relative norm, but change $\rho$ by only $6.61\%$ at the first step and $1.33\%$ at step eight. The incoming messages become increasingly redundant over the rollout: their mean cosine similarity rises from $0.592$ to $0.924$, the leading singular mode accounts for $90.7\%$ and $98.9\%$ of message energy, and the shared component accounts for $65.3\%$ and $97.6\%$, respectively. Accordingly, the same permutations increase total BCE by only $0.407\%$ and $0.154\%$. Writing $\Psi_\delta=\bar{\Psi}+\varepsilon_\delta$ gives
\[
\rho^\pi-\rho
=
\sum_\delta
\bigl(D_{\pi(\delta)}-D_\delta\bigr)\varepsilon_\delta ,
\]
because a permutation preserves $\sum_\delta D_\delta$ and therefore leaves the shared message component unchanged, explaining why a large parameter change can have a small functional effect. Step-eight geometry is measured on the model's own rollout states.

The cardinal preference nevertheless matters for expected excess BCE. For the full-sweep oracle $q_i(x)$ in Eq.~\eqref{eq:ising_full_sweep_oracle} and model probability $r_i(x)$, the conditional expected sitewise loss is
\begin{equation}
\mathbb E[\ell_{\rm BCE}(r_i,Z_i)\mid X=x]
=H_{\rm bin}(q_i)+\mathrm{KL}\!\left(
\operatorname{Ber}(q_i)\,\|\,\operatorname{Ber}(r_i)\right),
\label{eq:ising_expected_excess}
\end{equation}
where the BCE target is $(Z_i+1)/2$. The percentages below refer to the averaged KL term on the fixed diagnostic inputs, not subtraction of two realized test losses. This separates changes in the learned predictor from the irreducible conditional entropy. Setting all relation weights to their mean increases this excess loss by $34.5\%$; random permutations increase it by $67.5\%$ on average; and swapping the cardinal and diagonal weight groups increases it by $117.6\%$. All $40$ model--permutation pairs (eight fixed random permutations per seed) increase the excess loss, whereas within-cardinal or within-diagonal permutations have negligible effect. The cardinal--diagonal distinction is thus predictively meaningful, while exact assignment among redundant routes matters less. Zeroing $D$ is qualitatively different: it removes the relational signal altogether and raises the eight-step BCE from approximately $0.381$ to $0.699$.

\section{Evaluation Protocols and Additional Details}
\label{app:protocol}

\paragraph{Training and integration.} Continuous benchmarks use Smooth $L_1$ losses and binary lattices use cross-entropy; increment targets, where used, are finite observation differences. Checkpoints are selected by validation performance. Continuous benchmarks use Heun updates, recomputing emissions, aggregates, and responses at both stages with $D$ fixed; learned model time need not numerically equal physical observation spacing. Gradients reach $D$ through the rollout without a thresholded graph, and regularizers act on magnitudes rather than designated neighbours.

Table~\ref{tab:hyperparameters} summarizes the main AugRelNet configurations across benchmarks; benchmark-specific data and evaluation protocols are given below.
\begin{table}[t]
\centering
\setlength{\tabcolsep}{3.5pt}
\caption{Main AugRelNet configurations across benchmarks. Unless stated otherwise,
models use AdamW with initial learning rate $8\times10^{-4}$, weight decay
$10^{-4}$, gradient clipping at $1$, SiLU activations, and validation-based
checkpoint selection.}
\label{tab:hyperparameters}
\begin{tabular}{lccccc}
\toprule
Setting & GLV & Lorenz--96 & FHN & Life & Ising \\
\midrule
History length
& 32 & 32 & 2 & 9 & 5 \\
$(d_u,d_\eta)$
& $(8,8)$ & $(96,96)$ & $(64,64)$ & $(8,8)$ & $(8,8)$ \\
Message dimension
& 8 & 96 & 128 & 16 & 16 \\
Conditioner dimension
& 8 & 96 & -- & 16 & 16 \\
Source MLP width(s)
& 32 & 512 & 384,384 & 64 & 64 \\
Dynamics MLP width(s)
& 48 & 576 & 384,384 & 64 & 64 \\
Candidate relations
& \shortstack{all\\non-self} & \shortstack{all\\non-self}
& \shortstack{24\\radius-2} & \shortstack{24\\radius-2} & \shortstack{8\\surrounding} \\
Update
& Heun & Heun & Heun & discrete & discrete \\
Training horizons
& \shortstack{1,4,8,16\\24,48,96,256}
& \shortstack{1,4,8,16\\24,40}
& 1,2,4,8,16
& 1,2,4,8,16
& 1,2,4,8 \\
Prediction loss
& Smooth $L_1$ & Smooth $L_1$ & Smooth $L_1$ & BCE & BCE \\
Batch size
& 128 & 128 & 4 & 2 & 8 \\
Explicit $D$ regularizer
& \shortstack{$10^{-5}$\\$\operatorname{mean}(D^2)$}
& \shortstack{$10^{-5}$\\$\operatorname{mean}(D^2)$}
& \shortstack{$2\times10^{-5}$\\$\operatorname{mean}|D|$}
& \shortstack{$10^{-5}$\\$\operatorname{mean}(D^2)$}
& \shortstack{$10^{-5}$\\$\operatorname{mean}(D^2)$} \\
Learning-rate schedule
& constant & \shortstack{constant, then\\halvings} & plateau & plateau & constant \\
\bottomrule
\end{tabular}
\end{table}

\subsection{GLV protocol and additional diagnostics}
\label{app:glv_diagnostics}
We use five independently generated six-species systems, each with 16 training, 4 validation, and 4 test trajectories of 1,200 observations at spacing $\Delta=0.035$. Classical RK4 integrates the dynamics in log-abundance coordinates with 32 substeps per observation for training and validation and 64 for testing. Each species is standardized using training data only; models receive a history of 32 observations. Evaluation uses 3,652 held-out origins per system at horizons $h\in\{16,96,256\}$. Both RMSE and $R^2$ in Table~\ref{tab:glv_prediction} are computed in standardized log-abundance coordinates, not raw abundance coordinates; reported values are means over the five systems.

In the four strongly aligned systems, absolute off-diagonal correlations range from $0.9951$ to $0.9967$. In the fifth, correlations for weak, intermediate, and strong edges are $0.056$, $0.902$, and $0.915$, respectively, alongside overall correlation $0.767$ and longest-horizon $R^2=0.995$. These observations motivate the target-dependent alignment analysis below.
\begin{table}[htbp]
\centering
\setlength{\tabcolsep}{3.5pt}
\caption{GLV held-out forecasting: means over five independently
generated systems.}
\label{tab:glv_prediction}
\begin{tabular}{llccc}
\toprule
Method & Metric & $h=16$ & $h=96$ & $h=256$\\
\midrule
\multirow{2}{*}{Neural ODE}
& Standardized RMSE & 0.1497 & 0.3504 & 0.4724\\
& $R^2$ & 0.9840 & 0.9029 & 0.7902\\
\multirow{2}{*}{AugRelNet}
& Standardized RMSE & \textbf{0.0294} & \textbf{0.0607} & \textbf{0.1870}\\
& $R^2$ & \textbf{0.9991} & \textbf{0.9960} & \textbf{0.9466}\\
\bottomrule
\end{tabular}
\end{table}

\subsection{Lorenz--96}

\paragraph{Data and evaluation.} The main experiment uses 3,849 consecutive training states at observation spacing $\Delta=0.02$, followed by validation data and 785 held-out test origins. Normalization is fitted on the training segment only. Main-table RMSE is reported in physical coordinates; intervention scores explicitly labelled as standardized use normalized coordinates. Unless stated otherwise, results summarize five model seeds.

\paragraph{Optimization and diagnostic cohorts.} The main AugRelNet row uses validation-best checkpoints after an initial 3,500 epochs followed by 12,988--16,347 additional epochs of lower-learning-rate optimization across seeds. Continuation starts at $4\times10^{-4}$ and uses three validation-triggered halvings to a final backbone rate of $5\times10^{-5}$. Rolling-$b$ and intervention diagnostics use the initial pre-continuation checkpoints. State-ablation controls are independently validation-selected.

\paragraph{Baselines.} We compare with MNO~\citep{li2022chaotic}, DySLIM~\citep{schiff2024dyslim}, DeepSkip~\citep{mandal2025random}, AL-RNN~\citep{brenner2024almost}, a global echo-state reservoir based on \citet{platt2022systematic}, Neural ODE~\citep{chen2018neural}, LSTM~\citep{hochreiter1997long}, iTransformer~\citep{liu2024itransformer}, and a forecasting adaptation of NRI~\citep{kipf2018nri}. History-based models use the same 32-observation causal context where required. NRI infers a deterministic MAP graph from the input history and holds it fixed during autoregressive rollout.

GraphODE-Dense uses all non-self edges, whereas GraphODE-Oracle receives the 120 true directed edges without their signed coefficients. PySINDy~\citep{kaptanoglu2022pysindy} is evaluated with quadratic and cubic libraries; the quadratic library matches the generating equation class. All controlled comparisons use the same low-data split. For context, representative published 40-dimensional Lorenz--96 studies use about $3.8\times10^4$ training samples for local reservoir computing \citep{platt2022systematic} and $10^5$ for DeepSkip \citep{mandal2025random}.
\begin{table}[htbp]
\centering
\setlength{\tabcolsep}{4pt}
\caption{Full Lorenz--96 baseline comparison under the common low-data
protocol. Physical RMSE on 785 held-out origins. Mean $\pm$ population
standard deviation over five model seeds (PySINDy:
deterministic).}
\label{tab:l96_baseline_full}
\begin{tabular}{lcccc}
\toprule
Method & $h=1$ & $h=8$ & $h=24$ & $h=40$ \\
\midrule
\multicolumn{5}{l}{\textit{Dynamical-system models}} \\
MNO
& $0.768\pm0.047$
& $2.675\pm0.057$
& $4.362\pm0.117$
& $4.582\pm0.069$ \\
DySLIM
& $0.394\pm0.007$
& $2.919\pm0.061$
& $8.301\pm0.188$
& NF (4/5) \\
Global DeepSkip
& $0.151\pm0.003$
& $1.126\pm0.027$
& $3.627\pm0.061$
& $14.918\pm18.298^{\dagger}$ \\
AL-RNN
& $0.642\pm0.022$
& $2.449\pm0.015$
& $3.512\pm0.055$
& $3.687\pm0.033$ \\
Global ESN
& $0.440\pm0.004$
& $2.218\pm0.007$
& $3.488\pm0.009$
& $3.441\pm0.010$ \\
\midrule

\multicolumn{5}{l}{\textit{General forecasting models}} \\
Neural ODE
& $0.355\pm0.001$
& $2.341\pm0.006$
& $3.364\pm0.010$
& $3.362\pm0.007$ \\
Global LSTM
& $0.278\pm0.002$
& $2.010\pm0.012$
& $3.415\pm0.023$
& $3.555\pm0.024$ \\
iTransformer
& $0.346\pm0.019$
& $1.591\pm0.018$
& $4.199\pm0.008$
& $4.274\pm0.005$ \\
\midrule

\multicolumn{5}{l}{\textit{Relation-aware models}} \\
\shortstack[l]{NRI (forecasting\\adaptation)}
& $0.114\pm0.012$
& $0.710\pm0.056$
& $3.888\pm0.118$
& $5.082\pm0.100$ \\
GraphODE-Dense
& $0.378\pm0.001$
& $2.711\pm0.010$
& $4.857\pm0.043$
& $5.014\pm0.065$ \\
\midrule

\multicolumn{5}{l}{\textit{AugRelNet}} \\
AugRelNet (fixed-$b$)
& $\mathbf{0.0061}\pm0.0003$
& $\mathbf{0.049}\pm0.003$
& $\mathbf{0.250}\pm0.016$
& $\mathbf{0.685}\pm0.051$ \\
\shortstack[l]{AugRelNet (fixed-$b$)\\(initial checkpoints)}
& $0.014\pm0.001$
& $0.083\pm0.005$
& $0.333\pm0.020$
& $0.833\pm0.048$ \\
\shortstack[l]{AugRelNet (rolling-$b$)\\(initial checkpoints)}
& $0.014\pm0.001$
& $0.293\pm0.040$
& $1.103\pm0.142$
& $2.181\pm0.305$ \\
\midrule

\multicolumn{5}{l}{\textit{Strong-prior references}} \\
\shortstack[l]{GraphODE-Oracle\\(true graph)}
& $0.007\pm0.001$
& $0.055\pm0.007$
& $0.203\pm0.039$
& $0.451\pm0.083$ \\
\shortstack[l]{PySINDy poly2\\(quadratic class)}
& $0.010$
& $0.078$
& $0.271$
& $0.576$ \\
\midrule

\multicolumn{5}{l}{\textit{Additional dictionary-fit diagnostic}} \\
\shortstack[l]{PySINDy poly3\\(zero-field fit)$^{\ddagger}$}
& $0.390$
& $2.859$
& $5.384$
& $5.708$ \\
\bottomrule
\end{tabular}

\vspace{3pt}
\begin{minipage}{0.97\textwidth}
NF: four DySLIM seeds are nonfinite at $h=40$; no finite-subset mean.
$^\dagger$ All five DeepSkip runs are finite; the $51.514$ outlier is
retained.
$^\ddagger$ Poly3's validation-selected zero field gives persistence.
GraphODE-Oracle receives the true 120 directed edges without their signed
coefficients; PySINDy poly2 uses the matched quadratic dictionary.
\end{minipage}
\end{table}

\paragraph{Structure and interventions.} Global top-120 recovery ranks all non-self $|D_{ij}|$ entries; exact recovery means that these 120 entries coincide with the full true support. For NRI, structure scores are obtained from edge-existence probabilities averaged over the test histories. For the extended AugRelNet models, all five top-120 counts are 120 and all 40 signed pairs are opposed in every seed. Edge/nonedge mean-absolute-weight ratios are $153.98$, $84.45$, $304.84$, $108.60$, and $78.79$ (seed order 0--4); their arithmetic mean is $146.13$. The minimum-true-edge minus maximum-nonedge margins are $0.10695$, $0.08249$, $0.06988$, $0.10817$, and $0.07980$, respectively. Figure~\ref{fig:l96_relations_spacing}(b) uses the extended seed-0 checkpoint.

The $D$ interventions either zero all relation weights or shuffle their off-diagonal assignments while preserving the weight multiset. The $\eta$-reset diagnostic uses 32 fixed test origins, completes each Heun step normally, and then resets only $\eta$ to its initial value; the corrector therefore still uses the predicted response state within that step. State-factor ablations retain the relation pathway and history conditioner while adjusting local widths so that trainable parameter counts remain matched within approximately $0.2\%$.

The no-direct-shrinkage diagnostic removes both the explicit $D$ penalty and AdamW decay on $D$ in a seed-0 run; the remaining protocol is unchanged. The coarser-observation study uses three seeds each at $\Delta=0.05$ and $0.10$, with 2,200 observations and the same architecture in both settings. A two-hop shell contains directed pairs whose shortest path in the true interaction graph is exactly two; shell energy is the sum of squared learned weights normalized by total off-diagonal squared weight.

\paragraph{Capacity-balanced state ablations.}
\begin{table}[htbp]
\centering
\setlength{\tabcolsep}{3pt}
\caption{Capacity-balanced Lorenz--96 state ablations.
Physical RMSE over five seeds; all rows retain a trainable $D$.
$u$ is freely evolving augmentation, whereas $\eta$ uses the zero-anchored
relational drive. The full-state row is the initial-checkpoint reference.
Counts are trainable parameters in millions.}
\label{tab:l96_state_ablations}
\begin{tabular}{lccc}
\toprule
Evolving state & Params & $h=8$ & $h=40$\\
\midrule
$z,u,\eta$ & 1.2927 & $\mathbf{.083}\pm.005$ & $\mathbf{.833}\pm.048$\\
$z,\eta$ & 1.2934 & $.107\pm.005$ & $.894\pm.029$\\
$z,u$ & 1.2953 & $.131\pm.011$ & $1.120\pm.014$\\
$z$ only & 1.2950 & $.367\pm.020$ & $1.853\pm.057$\\
\bottomrule
\end{tabular}
\end{table}

The $(z,u)$ row is a capacity-matched free-augmentation control: it tests whether the gain from $\eta$ is explained by additional evolving capacity. Its higher error than $(z,\eta)$ supports the response-state parameterization at matched capacity.

\paragraph{Additional diagnostic scores.}
\label{app:l96_additional_diagnostics}
An additional seed-0 run without direct shrinkage on $D$ has physical RMSEs $0.019$, $0.095$, $0.341$, and $0.843$ at $h=1,8,24,40$, while retaining strict support separation and the opposed signed pair at every target.

\subsubsection{Data scaling and long-rollout diagnostics}
\label{app:l96_scaling_longrun}

\paragraph{Controlled data scaling.} The scaling extension uses nested prefixes of one training trajectory with 1,000, 3,849, 10,000, and 50,000 complete states. Validation and testing use common origins across budgets, and each budget uses one model seed; this extension therefore complements rather than replaces the main five-seed 3,849-state comparison.

\begin{table}[htbp]
\centering
\setlength{\tabcolsep}{3.2pt}
\caption{Lorenz--96 physical RMSE in the controlled data-scaling extension.
Lower is better. One model seed is used per training budget.}
\label{tab:l96_scaling_full}
\begin{tabular}{rlrrrr}
\toprule
Training states & Model & $h=1$ & $h=8$ & $h=24$ & $h=40$ \\
\midrule
1,000 & AugRelNet & 0.03988 & 1.1830 & 3.0405 & 3.6249 \\
      & DeepSkip  & 0.48823 & 3.4118 & 728.623 & 4196.22 \\
3,849 & AugRelNet & 0.00696 & 0.05269 & 0.27787 & 0.70441 \\
      & DeepSkip  & 0.15142 & 1.13883 & 3.76686 & 5.58873 \\
10,000 & AugRelNet & 0.00734 & 0.03002 & 0.11544 & 0.35828 \\
       & DeepSkip  & 0.04035 & 0.30323 & 1.03507 & 2.05554 \\
50,000 & AugRelNet & 0.00512 & 0.01476 & 0.03543 & 0.07051 \\
       & DeepSkip  & 0.00888 & 0.06587 & 0.22108 & 0.48991 \\
\bottomrule
\end{tabular}
\end{table}

\begin{table}[htbp]
\centering
\setlength{\tabcolsep}{5pt}
\caption{AugRelNet relation recovery across the same scaling budgets.
The separation margin is the smallest true-edge magnitude minus the
largest nonedge magnitude; positive values imply strict global separation.}
\label{tab:l96_scaling_structure}
\begin{tabular}{rrrr}
\toprule
Training states & \shortstack{Top-120\\true edges} & \shortstack{Separation\\margin} & \shortstack{Edge/nonedge\\contrast} \\
\midrule
1,000  & 104/120 & -0.0997 & 5.35 \\
3,849  & 120/120 &  0.1110 & 108.33 \\
10,000 & 120/120 &  0.1190 & 157.14 \\
50,000 & 120/120 &  0.1767 & 168.95 \\
\bottomrule
\end{tabular}
\end{table}

\paragraph{Long autonomous rollouts.} For the long-rollout diagnostics, the locked 50k-trained AugRelNet is run autonomously for 50,000 steps from five 32-state initial histories. The first 5,000 steps are excluded from post-burn-in statistics; augmented states evolve continuously, with no truth refresh or hidden-state reset, and the existing observable clamp is never activated. DeepSkip is included where the corresponding long-rollout comparison was computed.

\begin{table}[htbp]
\centering
\setlength{\tabcolsep}{6pt}
\caption{Selected post-burn-in Lorenz--96 long-run statistics. For mean
energy, the final column is the reference truth value; for discrepancy
metrics it is an independent-truth versus truth finite-sample reference,
not a significance threshold. Lower is better for discrepancy metrics.}
\label{tab:l96_longrun_statistics}
\begin{tabular}{lrrr}
\toprule
Metric & AugRelNet & DeepSkip & Reference \\
\midrule
Mean energy per site & 12.6763 & 9.3832 & 9.3618 \\
Marginal $W_1/\sigma$ $\downarrow$ & 0.23577 & 0.00183 & 0.00211 \\
Ordered ACF $L_2$ $\downarrow$ & 0.04010 & 0.00232 & 0.00416 \\
Temporal-spectrum JS $\downarrow$ & 0.007220 & 0.000234 & 0.000319 \\
\bottomrule
\end{tabular}
\end{table}

\begin{figure}[htbp]
    \centering
    \includegraphics[width=\linewidth]{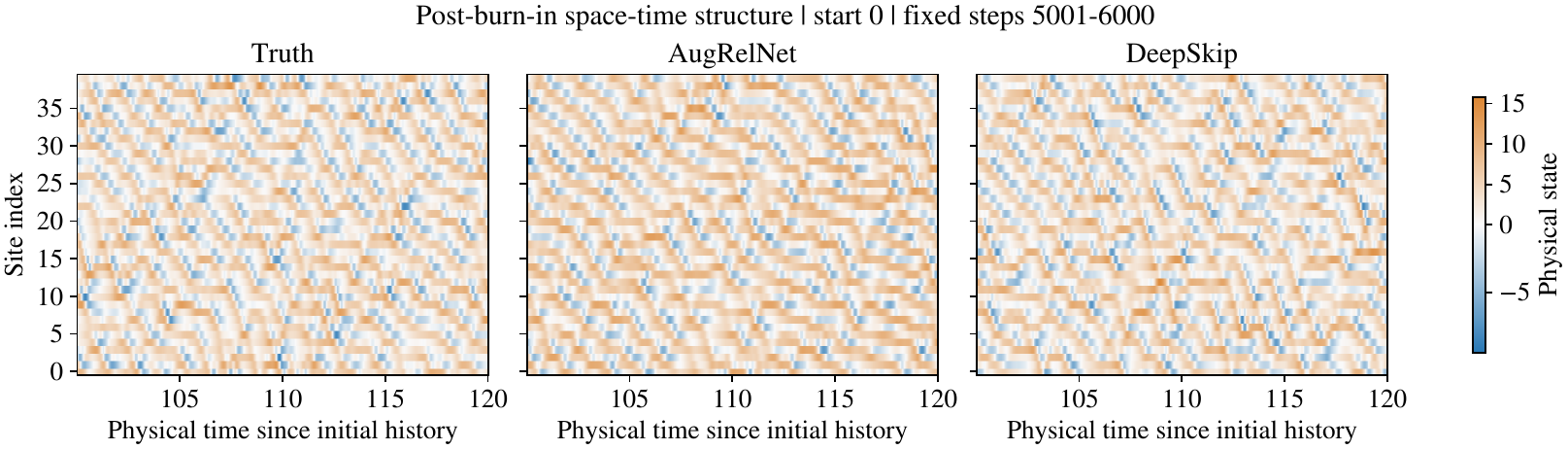}
    \caption{\textbf{Post-burn-in Lorenz--96 space--time structure.}
    Reference dynamics, AugRelNet, and DeepSkip for fixed steps
    5001--6000 from start 0. This is a qualitative long-rollout diagnostic;
    pointwise phase agreement is not expected after chaotic divergence.}
    \label{fig:l96_long_spacetime}
\end{figure}

\begin{figure}[htbp]
    \centering
    \includegraphics[width=\linewidth]{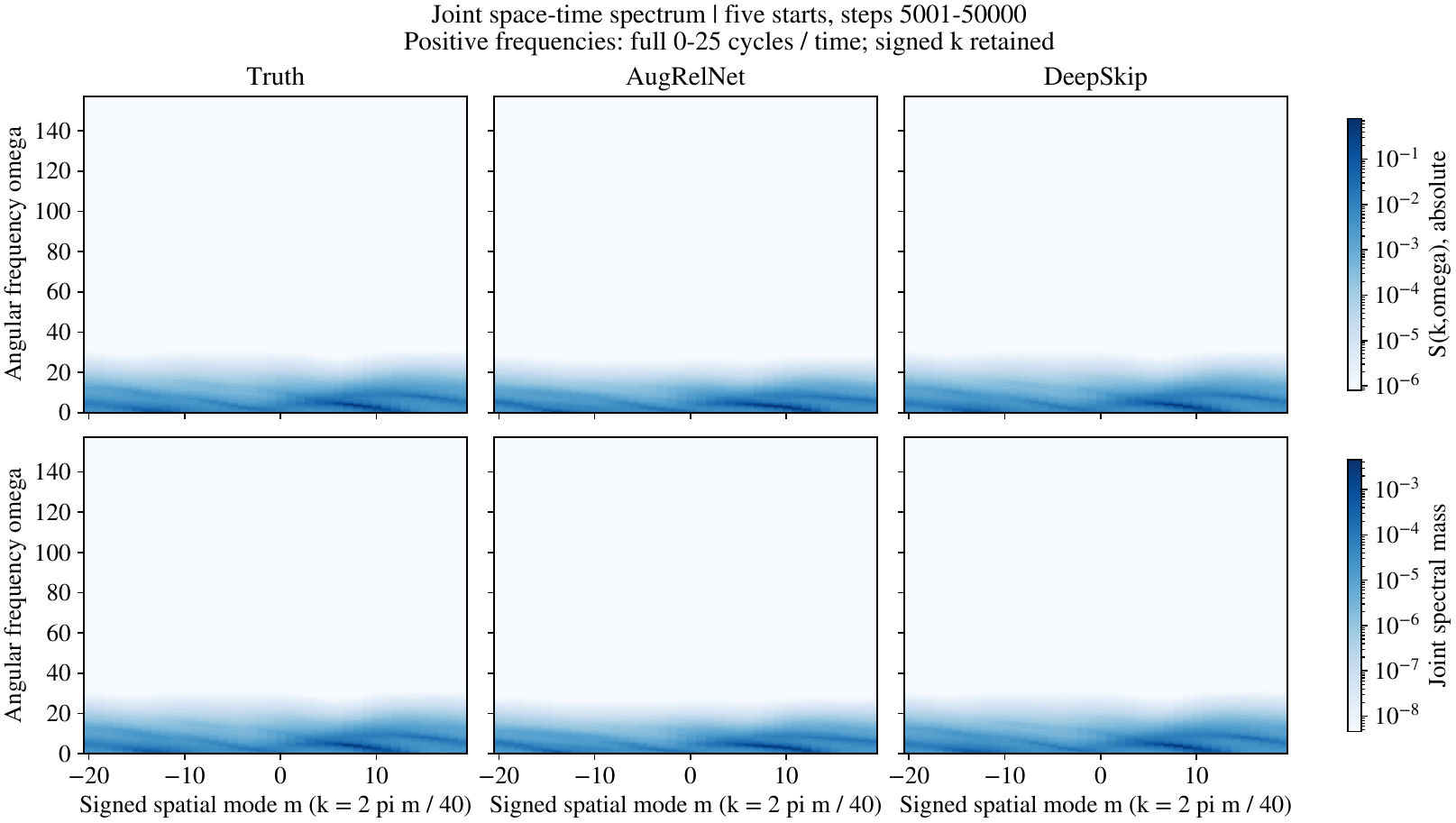}
    \caption{\textbf{Joint space--time spectra after burn-in.}
    Spectra are aggregated over five starts and steps 5001--50000.
    The displayed frequency window is fixed across reference dynamics,
    AugRelNet, and DeepSkip.}
    \label{fig:l96_joint_spacetime_spectrum}
\end{figure}

\begin{figure}[htbp]
    \centering
    \includegraphics[width=\linewidth]{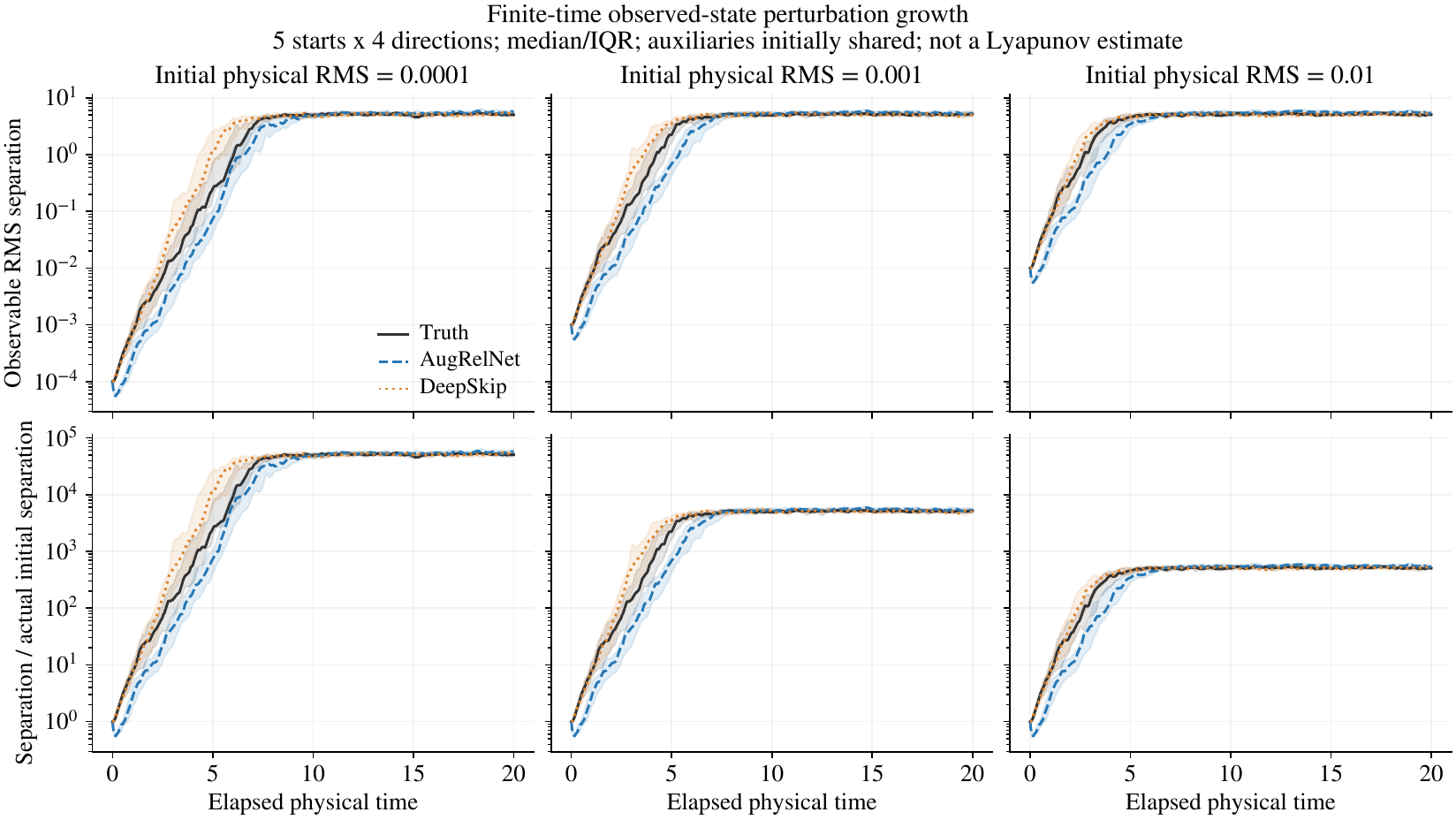}
    \caption{\textbf{Finite-time observed-state perturbation growth.}
    Five starts and four perturbation directions are evaluated at three
    initial physical RMS scales; curves show the median and interquartile
    range. Auxiliary states are initially shared. This diagnostic is not a
    Lyapunov-exponent estimate.}
    \label{fig:l96_perturbation_growth}
\end{figure}

\begin{figure}[htbp]
    \centering
    \includegraphics[width=\linewidth]{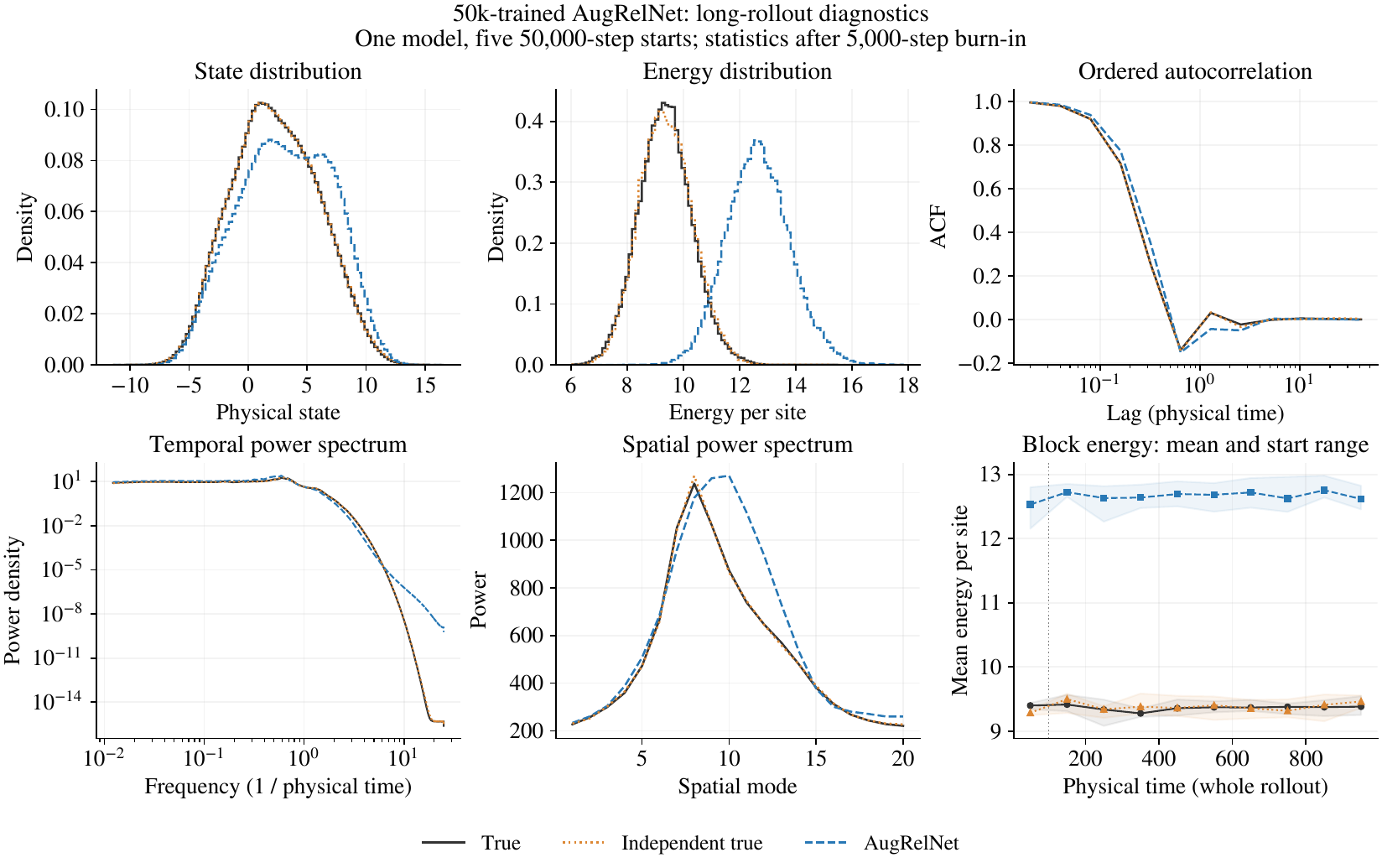}
    \caption{\textbf{Long-rollout diagnostics for the 50k-trained AugRelNet.}
    Five 50,000-step starts are summarized after a 5,000-step burn-in.
    Panels report state and energy distributions, ordered autocorrelation,
    temporal and spatial power spectra, and blockwise mean energy; the
    independent-truth curve or range provides a finite-sample reference.}
    \label{fig:l96_long_statistics}
\end{figure}

\subsection{Lattice Implementations and Evaluation}

\paragraph{Recorded-step writeback.}
\label{app:lattice_writeback}
After a local augmented update $\widetilde q_p^{n+1}=\mathcal U_\theta(q^n,b;D)_p$, the visible prediction and retained auxiliaries are written back as
\begin{equation}
q_p^{n+1}=
\bigl[\mathcal C_\theta(\widetilde q_p^{n+1}),
      \Pi_{\rm aux}\widetilde q_p^{n+1}\bigr].
\label{eq:lattice_writeback}
\end{equation}
FHN uses $\mathcal C_\theta=\Pi_z$ and Heun integration. Life writes occupancy probabilities in $[0,1]$, while Ising writes expected spins in $[-1,1]$; these values and the retained auxiliaries supply subsequent emissions without future observations.

For the lattice realization in Eq.~\eqref{eq:lattice_aggregation}, translation sharing is an explicit symmetry prior: relation coordinates are indexed by candidate offsets rather than site pairs, so their count is independent of lattice area for a fixed stencil. FHN uses no offset descriptor and sets $c_{\mathcal N}=|\mathcal N|^{-1/2}$; the binary realizations use offset-conditioned source maps and $c_{\mathcal N}=1$.

\paragraph{FitzHugh--Nagumo.} The two-field FHN dataset is defined on a $64\times64$ grid with homogeneous Neumann boundaries. We use 20 training trajectories of 501 recorded frames at spacing $\Delta=0.01$; the underlying simulator uses RK4 with step $0.001$. AugRelNet, U-Net, and FNO use the same two input frames, train/validation arrays, train-only scalers, training horizons $h=1,2,4,8,16$, and equally weighted state/increment Smooth $L_1$ losses. Evaluation uses 512 common test origins and reports standardized RMSE in Table~\ref{tab:lattice_forecasts}. AugRelNet considers the 24 noncentral radius-two offsets and masks invalid boundary sources.

The FHN visible and lifted dynamics use a centered relational response, whereas its response drive uses $G(q,\rho)$ directly. The relaxation rate is parameterized as
\[
\lambda=\operatorname{sigmoid}(\widetilde{\lambda})/\Delta .
\]
Prediction and finite-increment Smooth $L_1$ terms receive equal weight. The AugRelNet model has about $1.02$M trainable parameters; U-Net has $7.76$M and FNO $0.93$M (counting complex weights as two real scalars). Final checkpoints are selected by validation performance.

\paragraph{Game of Life.} Life uses periodic $64\times64$ grids with 16 training trajectories of 512 frames. AugRelNet considers 24 noncentral radius-two offsets rather than prescribing the eight-neighbour Moore stencil. The CNN and PolyKAN references use the same main test arrays and forecast origins. PolyKAN is given a $3\times3$ convolutional receptive field but not the Life update rule. The separately generated confirmation set is evaluated without retraining.

\paragraph{Ising.} The Ising experiments use a zero-field ferromagnetic model with $J=1$, $\beta=0.4406867935$, periodic boundaries, and checkerboard heat-bath updates. Each recorded step is one full sweep with independently randomized sublattice order, not one site update; the oracle integrates out that order and intermediate neighbour spins as in Eq.~\eqref{eq:ising_full_sweep_oracle}. The dataset contains 64 training, 8 validation, and 8 test sequences of 256 frames. Five causal input frames yield 1,952 shared evaluation origins, and the candidate stencil contains the four cardinal and four diagonal offsets.

During autoregressive rollout, site logits are converted to expected spins $2\sigma(\mathrm{logit})-1$ and written back without sampling future spin configurations. Cross-entropy therefore evaluates sitewise predictive probabilities rather than the full joint transition law. Message-geometry and permutation diagnostics use 256 fixed test origins; headline forecasting and one-step oracle comparisons use all 1,952 evaluation origins.

\end{document}